%% file: main.tex
\documentclass[10pt,twocolumn,letterpaper]{article}

\usepackage[pagenumbers]{cvpr} 

\newcommand{\env}{VISTA-Gym\xspace}
\newcommand{\ours}{VISTA-R1\xspace}

\usepackage[utf8]{inputenc}
\usepackage[T1]{fontenc}
\usepackage{times}
\usepackage{latexsym}
\usepackage{inconsolata}

\usepackage{amsmath}
\usepackage{amsfonts}
\usepackage{amssymb}
\usepackage{amsthm}
\usepackage{mathtools}
\usepackage{mathrsfs}
\usepackage{bm}
\usepackage{bbm}
\usepackage{nicefrac}
\usepackage{dsfont}

\usepackage{booktabs}
\usepackage{multirow}
\usepackage{tabularx}
\usepackage{array}
\usepackage{colortbl}
\usepackage{threeparttable}
\usepackage{float}
\usepackage{graphicx}
\usepackage{wrapfig}
\usepackage{caption}
\usepackage{tcolorbox}
\usepackage{makecell}

\usepackage[dvipsnames]{xcolor}
\usepackage{transparent}

\usepackage{enumitem}
\usepackage{xspace}
\usepackage{pifont}
\usepackage{lipsum}
\usepackage{upquote}
\usepackage{titletoc}

\usepackage[ruled,vlined]{algorithm2e}
\usepackage{algpseudocode}

\usepackage{listings}

\usepackage[linewidth=0.5pt,leftmargin=1em,rightmargin=1em,
    innertopmargin=5pt,innerbottommargin=5pt,
    innerrightmargin=5pt,innerleftmargin=5pt]{mdframed}

\usepackage[toc,page,header]{appendix}

\usepackage{natbib}

\usepackage{url}

\usepackage{smiles}
\usepackage{macros}
\input{math_commands.tex}

\definecolor{cvprblue}{rgb}{0.21,0.49,0.74}
\usepackage[pagebackref,breaklinks,colorlinks,allcolors=cvprblue]{hyperref}

\def\paperID{17590} 
\def\confName{CVPR}
\def\confYear{2026}

\title{Scaling Agentic Reinforcement Learning for Tool-Integrated Reasoning in VLMs}

\author{\textbf{Meng Lu}$^1$\footnotemark[1], \textbf{Ran Xu}$^2$\thanks{Equal contribution first authors: Meng Lu, Ran Xu. $^\dag$Equal correspondence to: Di Jin, Wenqi Shi, Xuan Wang.}, \textbf{Yi Fang}$^1$, \textbf{Wenxuan Zhang}$^3$, \textbf{Yue Yu}$^4$, \textbf{Gaurav Srivastava}$^1$, \textbf{Yuchen Zhuang}$^4$,\\
\textbf{Mohamed Elhoseiny}$^3$, \textbf{Charles Fleming}$^5$, \textbf{Carl Yang}$^2$, \textbf{Zhengzhong Tu}$^6$, \textbf{Guanghua Xiao}$^7$,\\
\textbf{Yang Xie}$^7$, \textbf{Hanrui Wang}$^8$, \textbf{Di Jin}$^8$$^\dag$, \textbf{Wenqi Shi}$^7$$^\dag$, \textbf{Xuan Wang}$^1$$^\dag$\\
$^1$Virginia Tech $^2$Emory University $^3$KAUST $^4$Georgia Tech $^5$Cisco \\
$^6$TAMU $^7$UT Southwestern Medical Center $^8$Eigen AI\\
}

\begin{document}
\maketitle
\input{section/0-abs}    
\input{section/1-intro}
\input{section/2-related}
\input{section/3-method}

\input{section/4-exp}

\input{section/5-conclusion}

{
    \small
    \bibliographystyle{ieeenat_fullname}
    \bibliography{main}
}

\appendix
\input{section/6-appendix}

\end{document}

%% file: math_commands.tex
\def\eqref#1{equation~\ref{#1}}

\def\1{\bm{1}}

\DeclareMathAlphabet{\mathsfit}{\encodingdefault}{\sfdefault}{m}{sl}
\SetMathAlphabet{\mathsfit}{bold}{\encodingdefault}{\sfdefault}{bx}{n}



%% file: section/0-abs.tex
\begin{abstract}
While recent vision-language models (VLMs) demonstrate strong image understanding, their ability to ``think with images,'' i.e., to reason through multi-step visual interactions, remains limited. 
We introduce \env{}, a scalable training environment for incentivizing tool-integrated visual reasoning capabilities in VLMs. 
\env{} unifies diverse real-world multimodal reasoning tasks (7 tasks from 13 datasets in total) with a standardized interface for visual tools (\eg, grounding, parsing), executable interaction loops, verifiable feedback signals, and efficient trajectory logging, enabling visual agentic reinforcement learning at scale.
While recent VLMs exhibit strong text-only reasoning, both proprietary and open-source models still struggle with tool selection, invocation, and coordination. 
With \env{}, we train \ours{} to interleave tool-use with agentic reasoning via multi-turn trajectory sampling and end-to-end reinforcement learning. 
Extensive experiments across 11 public reasoning-intensive VQA benchmarks show that \ours{}-8B outperforms state-of-the-art baselines with similar sizes by 9.51\%-18.72\%, demonstrating \env{} as an effective training ground to unlock the tool-integrated reasoning capabilities for VLMs.
Code and data of \env{} and \ours{} can be found at \url{https://github.com/Lucanyc/VISTA-Gym}.
\end{abstract}

%% file: section/1-intro.tex
\section{Introduction}
\label{sec:intro}

\begin{figure*}[t]
    \centering
    \begin{subfigure}[t]{0.49\linewidth}
        \centering
        \includegraphics[width=\linewidth]{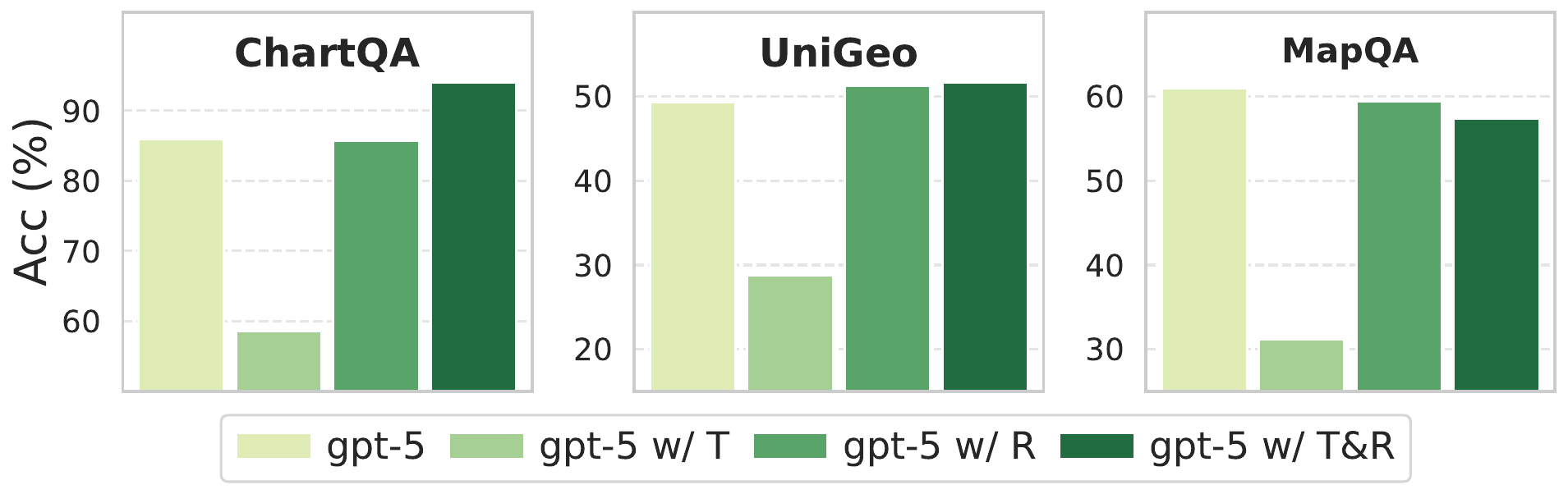}
        \caption{Close-source VLM (\texttt{gpt-5}).}
        \label{fig:teaser-gpt}
    \end{subfigure}
    \hfill
    \begin{subfigure}[t]{0.49\linewidth}
        \centering
        \includegraphics[width=\linewidth]{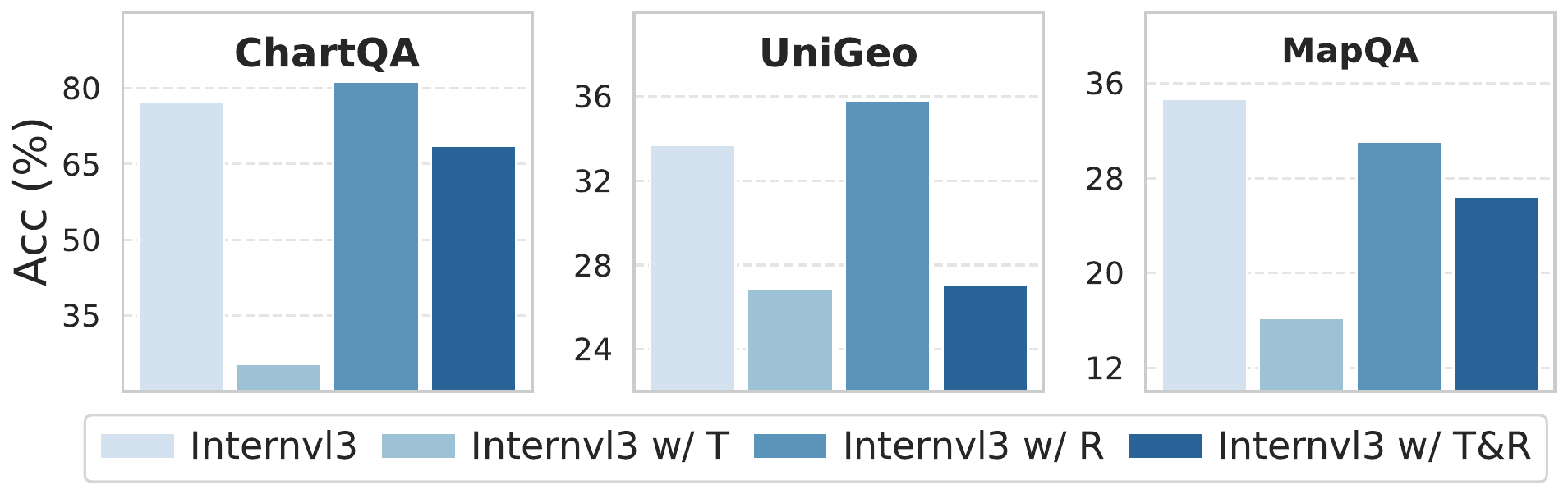}
        \caption{Open-source VLM (\texttt{Internvl3-8B}).}
        \label{fig:teaser-internvl}
    \end{subfigure}
    \caption{Directly augmenting VLMs with tools significantly degrades accuracy (\emph{w/ T}), yet intrinsic reasoning offers limited gains on complex VQA (\emph{w/ R}). Supplying \emph{tool‑selection prior knowledge} and interleaving reasoning with tool execution improve performance (\emph{w/ T\&R}); gains are task‑dependent for commercial VLMs, while small open‑source VLMs remain particularly struggling. }
    \label{fig:teaser}
\end{figure*}

Recent progress in VLMs has demonstrated strong performance across tasks such as visual question answering, multimodal reasoning, and grounded mathematical problem solving~\citep{bai2025qwen2,guo2025integrating,dong2025insight,thawakar2025llamav}. 
Many of these advancements stem from incorporating text-based reasoning paradigms, particularly the Chain-of-Thought (CoT)~\citep{wei2022chain} and reinforcement learning (RL)~\citep{guo2025deepseek}, which decomposes complex tasks into intermediate text reasoning steps for problem solving, and leverages outcome-based rewards to refine reasoning quality~\citep{liu2025visualrft,huang2025vision,qi2024cogcom,liu2024chain,zhang2025chain,shen2025zoomeye,liang2025modomodo,shen2025vgent}.

Most of the existing VLM reasoning processes still rely on static visual embeddings and shallow cross-modal alignment. As a result, their text-only reasoning struggles to capture the fine-grained visual structures, spatial relationships, and quantitative dependencies present in real-world scenes~\cite{zheng2025deepeyes, fan2025grit}.
These limitations underscore the need for thinking-with-image paradigms~\cite{su2025thinking}, where reasoning is tightly coupled with visual perception, enabling richer cross-modal interaction and step-by-step visual reasoning.

To enhance visual-centric reasoning and encourage \emph{thinking-with-image} behaviors in VLMs, tool-integrated reasoning (TIR)~\citep{guo2025beyond,zheng2025deepeyes,xu2025incentivizing} has been introduced recently. TIR equips models with external tools, such as \emph{grounding}, \emph{zoom-in}, and \emph{search}, to facilitate fine-grained perception and reasoning over object interactions.
Despite its promise, current open-sourced VLMs still remain inadequate at leveraging these tools for effective reasoning, as shown in Figure~\ref{fig:teaser}.
This limitation highlights the urgent need for an effective training environment and strategies to \emph{select, invoke, and coordinate} visual tools dynamically. 
While there are several studies dedicated to improving tool usage capabilities for VLMs~\citep{wu2025mmsearchr1,zheng2025deepeyes,fan2025grit,su2025openthinkimg}, those works are often narrow in scope, often confined to specific tasks (\eg, using image search for knowledge or zoom-in for object grounding). 
In parallel, recent efforts in gaming and robotics~\citep{chen2025g1,wang2025vagen} have introduced unified environments for training VLM agents. These platforms provide controllable dynamics, grounded feedback, and structured task spaces, offering clear advantages for robot learning and conceptually aligning with TIR. 
However, the simulation of tool-integrated thinking in VLMs—especially for open-domain visual reasoning—remains largely underexplored.

Motivated by these challenges, we introduce \env{} (\textbf{Vis}ual-centric \textbf{T}ool-integrated \textbf{A}gentic training environment), a scalable, agentic training environment designed to systematically enhance the reasoning and decision-making capabilities of VLM agents across complex, real-world scenarios. Specifically, \env{} wraps up visual tool operations and provides textual feedback on receiving tool call commands by VLMs, and features:
\begin{itemize}[leftmargin=*,noitemsep,topsep=2pt]
    \item \textbf{Diverse multi-modal reasoning tasks.} \env{} provides a comprehensive VQA suite spanning 7 reasoning tasks across 13 public datasets, supporting training and evaluation of agents with strong generalization and tool‑integration skills.  
    \item \textbf{Unified, extensible tool interface.} \env{} exposes a standardized API over 26 pre‑defined tools that support perception, symbolic manipulation, chart and document interpretation, grounding high‑level reasoning in structured intermediate results. 
    \item \textbf{Scalable interactive infrastructure.} \env{} accelerates agent training via multithreading, parallel execution, and sequential sampling, enabling efficient trajectory collection and large‑scale automated evaluation compatible with diverse agent scaffolds.  
\end{itemize}
Building on \env{}, we further develop \ours{}, a VLM-based agent trained for robust, tool-augmented reasoning. Across five in‑domain and six out‑of‑domain reasoning‑intensive VQA benchmarks, \ours{}‑8B surpasses state‑of‑the‑art open‑source baselines of comparable size by 9.51\%–18.72\%. \env{} proves to be an effective and scalable solution for developing VLM agents that can robustly interleave reasoning and tool use to solve complex, multi-step visual problems.

%% file: section/2-related.tex
\section{Related Works}
\label{sec:related}

\noindent\textbf{RL for VLM Reasoning.} 
Improving the reasoning capabilities for VLMs has been an active research front. 
\cite{xu2025llava} synthesize reasoning chains via distilling reasoning knowledge from teacher models. 
Inspired by the success of R1-style training~\citep{guo2025deepseek}, several works attempt to leverage RL \citep{huang2025vision,liu2025visualrft,shen2025vlm,liu2025visionreasoner,li2025self} to improve VLM reasoning capabilities on visual tasks, including general visual understanding and mathematical reasoning.

\noindent\textbf{RL for Tool-Integrated Reasoning in VLMs.}
To better characterize the visual information, recent works explore the ``thinking with images'' paradigm, which goes beyond standard reasoning steps by incorporating additional tools in the reasoning process~\cite{li2025tirbench,han2025tiger}. 
DeepMMSearch-R1~\citep{narayan2025deepmmsearchr1} and MMSearch-R1~\citep{wu2025mmsearchr1} leverage image search tools to augment the context with external knowledge. 
Other approaches~\citep{zheng2025deepeyes,fan2025grit,su2025openthinkimg,zhang2025chain} utilize \emph{zoom-in} operations to focus on fine-grained visual regions, thereby improving grounding and multi-step perception reasoning. 
ReLook~\citep{li2025relook} utilizes a VLM as an auxiliary tool to enable agentic training via cross-model interaction.
Despite these advances, most existing methods rely on a single specialized tool and are restricted to narrow task domains, thereby limiting their generalization to broader multimodal reasoning scenarios.

\noindent\textbf{RL Training Environment for Agentic Reasoning.}
To facilitate agentic model training, several frameworks have been proposed, including
GEM~\citep{liu2025gem}, 
AgentGym‑RL~\citep{xi2025agentgym}, 
SkyRL‑Gym~\citep{cao2025skyrl}, Collabrative GYM~\citep{shao2024collaborative} and RAGEN~\citep{wang2025ragen}
 for training general agents.
Beyond general frameworks, many efforts target domain-specific applications, such as text-based reasoning~\citep{stojanovski2025reasoninggym}, software engineering~\citep{swegym,yang2025swesmith,codegym,jain2025regym}, machine learning engineering~\citep{nathani2025mlgym,qiang2025mlesmith,qiang2025mledojo}, search and web browsing~\citep{xiong2025rag,xu2025acesearcher,chezelles2024browsergym}, and scientific reasoning~\citep{xu2025medagentgym,wang2022scienceworld}.
Moreover, the vast majority of these environments operate purely in text-only settings, providing limited support for multimodal grounding or visual reasoning. 
For VLMs, to the best of our knowledge, only two environments, namely VLM-Gym~\citep{chen2025g1} and VAGEN~\citep{wang2025vagen}, are designed to support VLM training, focusing respectively on compositional visual games and embodiment tasks that require intermediate state information. 
\env fills this gap by providing a scalable environment for tool-integrated RL, enabling systematic development of agentic visual reasoning for VLMs.

%% file: section/3-method.tex
\input{section/3-1-pre}
\input{section/3-2-env}

\input{section/3-3-method}

%% file: section/3-1-pre.tex
\section{Preliminaries}
\label{sec:pre}
\noindent \textbf{Problem Setup.} 
We study visually grounded question answering requiring complex reasoning over both image and text to derive a final prediction $\hat{y}$. 
The model is equipped with a predefined, finite tool set (\ie, action space) \(\mathcal{A}\). 
Formally, the objective is to synthesize a trajectory $U$ with interleaved reasoning steps $g_t$ and external tool invocations $a_t \in \mathcal{A}$. 
The resulting solution path is defined as $U=(g_0, a_0, \dots, g_T, \hat{y})$, where $T$ denotes the number of tool-interaction turns required to reach the final answer $\hat{y}$.



\noindent \textbf{Tool-Integrated Reasoning in VLMs.}
Formally, we frame TIR in VLMs as a Partially Observable Markov Decision Process (POMDP) defined by the tuple $\langle\mathcal{S},\mathcal{O},\mathcal{A},\mathcal{I},\mathcal{P},\mathcal{R}\rangle$, where
$\mathcal{I}$, $\mathcal{S}$, $\mathcal{O}$, $\mathcal{A}$, and $\mathcal{R}$ represent the spaces for instructions, environment states, agent observations, actions, and rewards, respectively.
Let $\mathcal{P}: \mathcal{S}\times\mathcal{A}\to\mathcal{S}$ represent the deterministic state transition function.
Following ReAct~\cite{yao2023react}, we structure the agent outputs to prioritize reasoning thought before action.
Given a problem description $x\in\mathcal{I}$, the history, and the current feedback, at each turn $t$, the agent generates the thought $g_{t+1}\sim\pi_\theta(\cdot|x,g_1,a_1,o_1,\cdots,g_t,a_t,o_t)$ first and the subsequent action $a_{t+1}\sim\pi_\theta(\cdot|x,g_1,a_1,o_1,\cdots,g_t,a_t,o_t,g_{t+1})$.
Following $a_{t+1}$, the environment transitions to a new state $s_{t+1}$ via $\mathcal{P}(s_{t+1}|s_t,a_t)$, and provides a new partial observation $o_{t+1}\sim\mathcal{O}(\cdot|s_{t+1})$.
Thus, the entire trajectory is:
\begin{equation*}
    \begin{aligned}
        \tau=(g_0,a_0,o_0,\cdots,o_{T-1},g_T,\hat{y})\sim\pi_\theta(\tau|x),
    \end{aligned}
\end{equation*}
where $\hat{y}$ is the final answer obtained from the model generations, and $\pi_\theta$ can be decomposed:
\begin{equation*}
    \begin{aligned}
        &\pi_\theta(\tau|x)=\pi_\theta(g_T|x,c_{T-1})\cdot\prod_{t=0}^{T-1}\pi_\theta(g_t,a_t|x,c_{t-1}) \\
        & =\pi_\theta(g_T|x,c_{T-1})\cdot\prod_{t=0}^{T-1}\pi_\theta(a_t|x,c_{t-1},g_t)\cdot\pi_\theta(g_t|x,c_{t-1}),
    \end{aligned}
\end{equation*}
where $c_{t-1}=(g_0,a_0,o_0,\cdots,g_{t-1},a_{t-1},o_{t-1})$ represents the interactive history up to $t-1$.

\noindent \textbf{Exploratory Experiments.}
\input{table/table-error-type}
We conduct exploratory experiments on three VQA tasks \cite{masry2022chartqa,chen2022unigeo,chang2022mapqa} with both proprietary and open-source VLMs.
As shown in Fig.~\ref{fig:teaser}, naively enabling tools for base models induces a sharp accuracy drop: \emph{without instruction or reasoning priors, external tools act as distractors rather than aids.} To diagnose failure modes, we annotate 500 tool-enabled errors across GPT‑5 and InternVL3‑8B (Table~\ref{tab:error-def}). 
The majority of failures arise from the \emph{if/when/which/how of tool calls}, schema/argument selection, and correctness (E1–E5); followed by \emph{incorrect post‑tool reasoning} (E6). These observations motivate us to train VLMs for TIR; To facilitate easy scale-up, we introduce \env{} (Figure~\ref{fig:overview}): we curate verifiable, visual‑centric tasks with tools (\cref{sec:4.1}), instantiate an interactive, executable environment (\cref{sec:4.2}) with scalable training facilities (\cref{sec:4.3}), and systematically improve open-source VLM agent within \env{} that \emph{interleaves} reasoning with tool execution (\cref{sec:training}).

%% file: table/table-error-type.tex
\begin{table}[t]
\centering
\renewcommand\arraystretch{0.93}
\caption{Error pattern identification and distribution from 500 error samples. Note that one case may contain multiple error types.}
\fontsize{8}{10}\selectfont\setlength{\tabcolsep}{0.3em}
\resizebox{1.0\linewidth}{!}{ %
\begin{tabular}{l @{\hskip2.5pt} | c | c | c }
\toprule
\textbf{ID} &  \textbf{Error Type} & \textbf{GPT-5} & \textbf{InternVL3-8B}\\
\midrule
E1 & Invocation schema violation (wrong function call structure) & 12.8\% & 3.8\%\\
E2 & Invalid argument name (wrong augment name) & 0.0\% & 44.8\%\\
E3 & Invalid argument value (wrong augment value format) & 18.2\% & 18.0\%\\
E4 & Incorrect argument value (wrong argument value content) & 39.6\% & 57.4\%\\
E5 & Invalid output from tool execution (wrong answer format)  & 25.6\% & 0.0\%\\
E6 & Incorrect reasoning from tool execution & 28.1\% & 64.8\% \\
\bottomrule
\end{tabular}
}
\label{tab:error-def}
\end{table}

%% file: section/3-2-env.tex
\section{\env: A Scalable Tool-Integrated Agentic Training Environment for VLMs}
\label{sec:env}

\begin{figure*}[t]
    \centering
    \includegraphics[width=\linewidth]{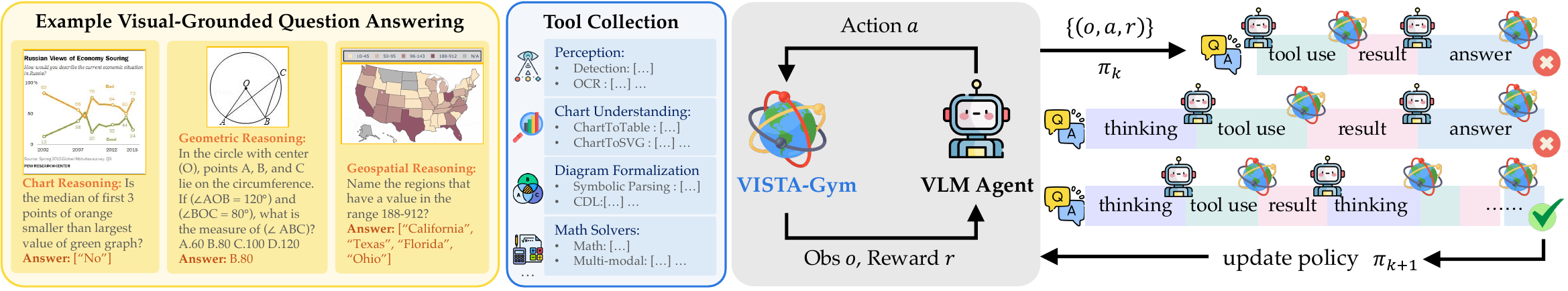}
    \caption{Overview of \env{}. \env{} contains a comprehensive suite of reasoning-intensive VQA tasks and tools in an interactive execution environment, scaling visual-centric tool-integrated agentic training for VLM agents. }
    \label{fig:overview}
\end{figure*}

\subsection{Diverse Task and Tool Collection} \label{sec:4.1}

\noindent\textbf{Reasoning-Intensive VQA Tasks.}
\env{} comprises a unified training and evaluation environment that couples diverse multimodal tasks with executable tools and verifiers, emphasizing not only final answers but also \emph{auditable sequences of tool calls} that ground those answers. We curate \emph{verifiable} instances from 13 established benchmarks to ensure broad coverage of difficulty and reasoning types, spanning perception (vision), symbolic manipulation (math/geometry), and language understanding (document/chart interpretation). Specifically, training tasks span seven complementary axes (Appendix~\ref{app:data}): 
(1) \emph{Chart Understanding} (FigureQA~\cite{ebrahimi2018figureqa}, ChartQA~\cite{masry2022chartqa}),
(2) \emph{Geometric Reasoning} (Geometry3K~\cite{lu2021inter}, GeoQA~\cite{chen2021geoqa}, UniGeo~\cite{chen2022unigeo}),
(3) \emph{Geospatial Reasoning} (MapQA~\cite{chang2022mapqa}, InfographicVQA~\cite{mathew2022infographicvqa}),
(4) \emph{Scientific Reasoning} (ScienceQA~\cite{lu2022learn}, VizWiz~\cite{bigham2010vizwiz}),
(5) \emph{Document Understanding} (DocVQA~\cite{mathew2021docvqa}),
(6) \emph{Spatial/Compositional Reasoning} (CLEVR~\cite{johnson2017clevr}),
and (7) \emph{Others} (ThinkVL~\cite{chitty2025imagenet}, A-OKVQA~\cite{schwenk2022okvqa}).

\noindent \textbf{Visual-Centric Tool Sets.}
Small open-source VLMs often lack the fine-grained perception and domain routines (\eg, precise localization, symbolic formalization, dense chart parsing) required by the above tasks. 
\env{} therefore exposes a standardized, extensible tool interface that lets agents offload these subproblems to reliable modules, \emph{grounding} high-level reasoning in structured intermediate results (\eg, tables, text). 
Twenty-six tools are organized into four families (Appendix~\ref{app:tool}): 
(1) \emph{Perception}, such as GroundingDINO~\cite{liu2023grounding}, SAM~\cite{kirillov2023segment}, EasyOCR;
(2) \emph{Chart Understanding}, such as ChartMoE~\cite{xu2025chartmoe};
(3) \emph{Diagram Formalization}, such as CDL, Inter‑GPS;
and (4) \emph{Math Solvers}, such as G‑LLaVA~\cite{gao2025gllava}, MultiMath~\cite{peng2024multimath}.

\subsection{Executable Interactive Environment} \label{sec:4.2}

\noindent \textbf{Interface.}
\env{} exposes a Gymnasium-style API with \texttt{reset()} and \texttt{step()}.
Given an instruction $x\in\mathcal{I}$, \texttt{reset} returns the initial partial observation $o_0$ (question text and supporting image[s]) and initializes the interaction history $c_0$.
Each episode is a POMDP mentioned in \cref{sec:pre}. 


\noindent \textbf{Action Space.}
The action space $\mathcal{A}$ is strictly constrained by the available toolset.
We formalize an action as a typed tuple $a_t \in \mathcal{A}$ consisting of the unique identifier and the corresponding arguments passed to the interface for execution.

\noindent \textbf{Observation Space.}
The observation space $\mathcal{O}$ comprises environmental feedback after tool execution.
Each $o_t\in\mathcal{O}$ encapsulates either successful execution results or runtime error messages (if any), serving as external signals for verifying intermediate hypotheses and adjusting the subsequent reasoning trajectory.


\subsection{Scalable Training Facility} \label{sec:4.3}
\noindent \textbf{VLMs as Tools.}
To enhance visual perception, we include compute‑intensive VLMs like {G-LLaVA}~\cite{gao2025gllava} and {ChartMoE}~\cite{xu2025chartmoe} as interactive services.
To integrate them efficiently into distributed RL, we deploy a highly concurrent microservice architecture that encapsulates each VLM as an independent HTTP service with three layers: 
(i) a \textbf{FastAPI} front end that exposes RESTful endpoints with asynchronous batched requests;
(ii) an intermediate \textbf{Tool} layer that parses instruction actions, retrieves image data from trajectory metadata, and formats observations; and
(iii) a \textbf{Ray Actor} layer that keeps model weights resident in GPU memory after initialization, eliminating the prohibitive latency of repeated model (re)loading under high‑frequency tool calls.


\noindent \textbf{Asynchronous Training.}
To sustain high throughput during RL rollouts, we use Ray to orchestrate concurrency. At each step, the policy emits a \texttt{<think>} segment followed by a \texttt{<tool\_call>}; on generation of the sentinel token \texttt{</tool\_call>}, decoding halts and the framework assembles batched HTTP requests that include trajectory identifiers and image paths. Ray manages request queues and load balancing across tool servers. For resource efficiency, compute‑heavy VLM tools are pinned to dedicated GPUs, while lightweight utilities are multiplexed on shared CPUs.  


\noindent \textbf{Extensible Tool Set.}
To support rapid expansion beyond tools used in our experiments, we provide a generic \texttt{BaseTool} interface that enables \emph{plug‑and‑play} integration of new tools with minimal boilerplate.
Operational robustness is ensured via health and metric endpoints (\texttt{/health}, \texttt{/metrics}) and Ray's automated failure recovery, which restarts crashed actors without disrupting ongoing training.


%% file: section/3-3-method.tex
\section{\ours: Bootstrapping Tool-Integrated Reasoning via RL for VLMs}
\label{sec:training}

\subsection{Two-Stage Training Framework}
With the developed \env, we enable a two-stage training framework. First, we bootstrap an interactive agent with behavioral cloning (BC), instilling instruction-following and basic tool-use. We then refine it with multi-turn, online RL, which encourages deeper reasoning and disciplined, reasoning-guided tool orchestration.

\begin{figure}
    \centering
    \includegraphics[width=\linewidth]{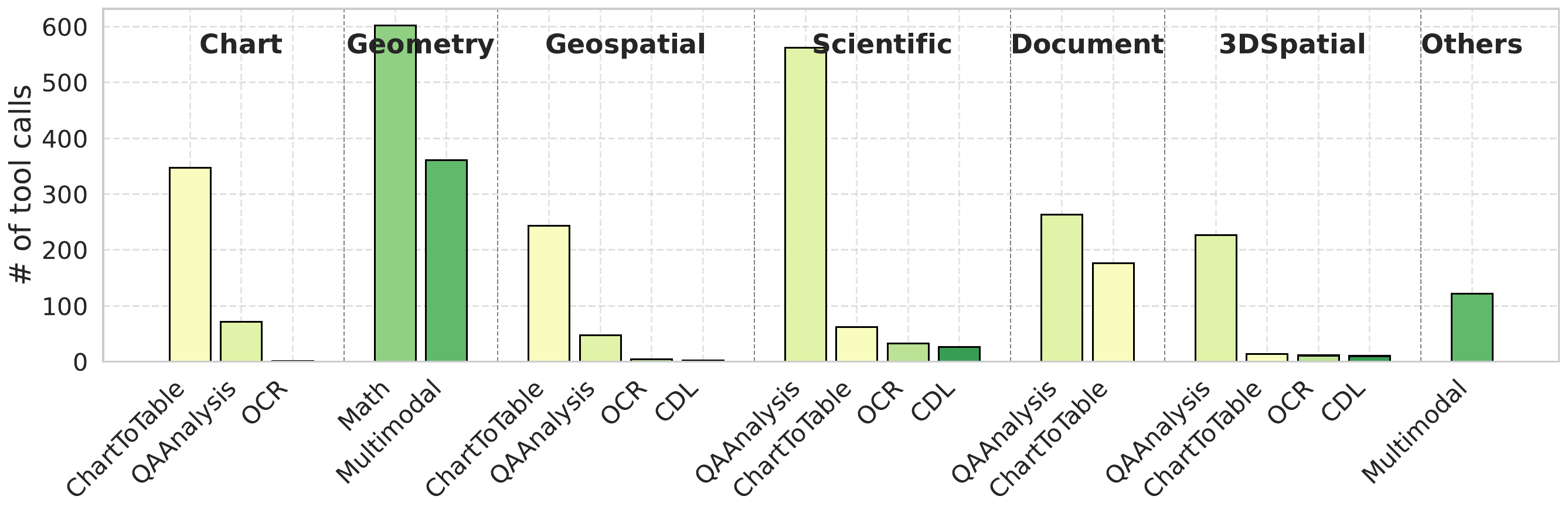}
    \caption{Top tool call distribution of different tasks in SFT data.}
    \label{fig:sft-tool-distribution}
\end{figure}

\noindent \textbf{Stage I: Warmup w. Imitation Learning.}
We initialize the policy by BC on synthesized expert trajectories that \emph{explicitly} interleave thoughts and actions. 
(i) \textit{Candidate generation and filtering.} We first generate tool‑executing trajectories with a proprietary model (\texttt{GPT‑5}) and retain only those whose final answers exactly match ground truth (outcome‑based filtering). 
(ii) \textit{Rationale densification.} To strengthen supervision on reasoning, we replace concise rationales with extended traces produced by an open‑weights expert (\texttt{Qwen3‑VL‑235B‑A22B‑Thinking}). 
Let $\mathcal{D}$ denote the resulting set of tuples $(x,\tau)$ where $\tau{=}(g_1,a_1,\ldots,g_T,\hat{y})$ and $c_{t-1}$ is the interaction history up to step $t{-}1$. The BC objective maximizes the likelihood of interleaved thought–action tokens:
\begin{equation*}
    \begin{aligned}
        &\mathcal{L}_{\text{BC}}(\theta)=\mathbb{E}_{(x,\tau)\sim\mathcal{D}}[\log\pi_\theta(\tau|x)]\\
        &=\mathbb{E}\left[\sum_{t=0}^{T-1}\log\pi_\theta(a_t|x,c_{t-1},g_t)+\sum_{t=0}^{T}\log\pi_\theta(g_t|x,c_{t-1})\right].
    \end{aligned}
\end{equation*}
The synthesized corpus covers a diverse tool mix (Figure~\ref{fig:sft-tool-distribution}), providing a robust prior for tool syntax and selection. 

\noindent \textbf{Stage II: Online RL.}
After SFT, we train the agent in the executable environment with multi-turn rollouts, where each step emits a reasoning segment followed by a function call; \env{} executes the call and appends the (structured) feedback to the context for continued reasoning.
For policy improvement, we adopt Group Relative Policy Optimization (GRPO)~\cite{shao2024deepseekmath} with group-normalized advantages over $G$ rollouts:
\begin{equation*}
    \begin{aligned}
        \mathcal{L}_{\text{GRPO}}(\theta)&=\frac{1}{G}\sum_{i=1}^G\frac{1}{|\tau_i|}\sum_{k=1}^{|\tau_i|}\min[r_{i,k}(\theta)\cdot\hat{A}_{i,k},\\
        &\quad\text{clip}(r_{i,k}(\theta),1-\epsilon,1+\epsilon)\cdot\hat{A}_{i,k}],
    \end{aligned}
\end{equation*}
where $r_{i,k}(\theta)$ denotes the token-level importance ratio at the $k$-th token.
\begin{equation*}
    \begin{aligned}
        r_{i,k}(\theta)&=\frac{\pi_\theta(\tau_{i,k}|\tau_{i,<k})}{\pi_{\text{old}}(\tau_{i,k}|\tau_{i,<k})},
    \end{aligned}
\end{equation*}
and $\hat{A}_{i,k}$ is the normalized advantage across all tokens:
\begin{equation*}
    \begin{aligned}
        \quad \hat{A}_{i,k}&=\frac{R(\tau_i)-\text{mean}(\{R(\tau_1),\cdots,R(\tau_G)\}}{\text{std}(\{R(\tau_1),\cdots,R(\tau_G)\})}.
    \end{aligned}
\end{equation*}

\subsection{Reward Design}
We employ a multi-round interaction protocol, $U=(u_0,u_1,\cdots,u_{T})$, for agent training, and design the following rewards $R$ for policy update. This protocol mandates the structural format for each turn $u_i$, enforcing an explicit \textit{think}$\rightarrow$\textit{tool\_call} loop prior to a \textit{think}$\rightarrow$\textit{answer} termination:

\noindent~$\bullet$ \textbf{Turn ($<T$) (Tool Call):} the first turns $u_{i<T}$ should generate reasoning with function calls: 
\begin{equation*}
    \begin{aligned} u_{i<T}=\text{<think>}\cdots\text{</think>}\text{<tool\_call>}\cdots\text{</tool\_call>}.
    \end{aligned}
\end{equation*}

\noindent~$\bullet$ \textbf{Turn $T$ (Final Answer):} the last turn $u_T$ should generate reasoning with final answer:
\begin{equation*}
    \begin{aligned}
u_T=\text{<think>}\cdots\text{</think>}\text{<answer>}\cdots\text{</answer>}.
    \end{aligned}
\end{equation*}

\noindent\textbf{Repetition Penalty.}
We first apply a high‑priority repetition detector $R_{\text{rep}}(U)\!\in\!\{-3.0,-2.0,-1.5,0\}$ that scans for contiguous token/phrase/character repeats and assigns a severity‑dependent negative reward (extreme, severe, moderate, none). This term dominates all subsequent logic. 


\noindent\textbf{Format Reward.}
Conditional on \emph{no} repetition, $R_{\text{rep}}(U){=}0$, we validate structural well‑formedness (\eg, correct tags, order, and non‑nested closure) at every turn. We define:
\begin{equation*}
    \begin{aligned}
        R_{\text{format}}(U)=(\mathbb{I}\{\text{all $u_i$ conforms the format}\}-0.5)\times2.
    \end{aligned}
\end{equation*}


\noindent\textbf{Correctness Reward.}
We extract the final prediction $\hat{y}$ from \texttt{<answer>}…\texttt{</answer>} for rule‑based checking. To keep the signal low‑noise and format‑aware, correctness is credited only for repetition‑free, well‑formed outputs:
\begin{equation*}
    \begin{aligned}
        R_{\text{correct}}(U)=\mathbb{I}\{\hat{y}=y\}.
    \end{aligned}
\end{equation*}

\noindent\textbf{Final Reward.}
The rollout‑level reward is the sum of the three components:
\begin{equation*}
    \begin{aligned}
        R(U)=R_{\text{rep}}(U)+R_{\text{format}}(U)+R_{\text{correct}}(U).
    \end{aligned}
\end{equation*}
This sparse, format‑aware design allocates positive reward only to generations that are repetition‑free, structurally valid, and correct, thereby encouraging the policy to internalize the intended \textit{think}$\rightarrow$\textit{tool\_call}$\rightarrow$\textit{answer} protocol rather than exploiting intermediate heuristics.  

%% file: section/4-exp.tex
\input{table/table-main}

\section{Experiments}
\label{sec:exp}

\subsection{Experiment Setups}
\noindent \textbf{Evaluation Datasets.}
Following several prior works \cite{liu2025visionreasoner,huang2025vision,han2025tiger,li2025self}, we focus on \emph{reasoning-intensive VQA tasks}, evaluating the effectiveness of \env{} in improving the TIR of open-source VLMs.
For \emph{in‑distribution} evaluation, we select five out of the thirteen datasets in \env{} training pool with official held-out test splits: (1) ChartQA~\cite{masry2022chartqa}, (2) Geometry3K~\cite{lu2021inter}, (3) GeoQA~\cite{chen2021geoqa}, (4) UniGeo~\cite{chen2022unigeo}, and (5) MapQA~\cite{chang2022mapqa}. 
To assess \emph{out‑of‑distribution} generalization, we evaluate on six additional benchmarks not used for training: (6) TABMWP~\cite{lu2022dynamic}, (7) AI2D~\cite{kembhavi2016diagram}, (8) PlotQA~\cite{methani2020plotqa}, (9) CLEVR-Math~\cite{lindstrom2022clevr}, (10) IconQA~\cite{lu2021iconqa}, and (11) MathVista~\cite{lu2023mathvista}.
Detailed dataset information is in Appendix \ref{app:data}.

\noindent \textbf{Baselines.}
We consider the following baselines on \env{}:
(i) \emph{API-based proprietary VLMs}, including GPT-5~\cite{gpt5}, GPT-5-mini~\cite{gpt5}, GPT-o4-mini~\cite{openai2025introducing}, GPT-o3~\cite{openai2025introducing}, Gemini-2.5-Pro~\cite{comanici2025gemini}, Gemini-2.5-Flash~\cite{comanici2025gemini}, Claude-4.5-Sonnet~\cite{Claude};
(ii) \emph{Open-source VLMs}, including InternVL3~\cite{zhu2025internvl3}, Qwen2.5-VL~\cite{bai2025qwen2}, LLaVA-OneVision-1.5~\cite{an2025llava};
(iii) \emph{Tool/Reasoning-integrated VLMs}, including VTool-R1~\cite{wu2025vtool}, R1-VL~\cite{zhang2025r1}, R1-Onevision~\cite{yang2025r1}, Perception-R1~\cite{yu2025perception}.
All baselines are evaluated \emph{without tool access}, as naive tool exposure significantly degrades performance without extensive agentic training (see Table~\ref{tab:fig-tir}); except for tool-integrated baselines following their original tool configurations.
Additional baseline details are available in Appendix~\ref{app:baseline}.

\noindent \textbf{Implementation Details.}
We consider four different backbones for \ours{} with varying sizes, including InternVL3-2B/8B/14B and Qwen2.5-VL-7B, and implement training with Verl-Tool~\cite{jiang2025verltool}. 
For SFT, we train for one epoch with a batch size of 128, a learning rate of $2\times10^{-6}$, and a context length of 16,384 tokens.
For RL, we use a micro-batch size of 8 per GPU, a mini-batch size of 128, and $G=8$ rollouts per update. We set the regularization coefficient $\beta=10^{-3}$, cap the maximum response length at 26,780 tokens, and optimize with a learning rate of $5\times10^{-7}$ for $E=300$ steps.
We adopt \emph{accuracy} (ACC) as the primary evaluation metric. All experiments are conducted with 8 NVIDIA H200 GPUs, each equipped with 141GB of memory.
See Appendix~\ref{app:prompt} and \ref{app:implementation} for prompt templates and other implementation details, respectively.

\subsection{Main Experiment Results}
Table~\ref{tab:main_results} shows the main results of \ours{} trained in \env{} against baselines on eleven visual‑reasoning datasets. From the results, we have the following key observations:
\textbf{(i) \ours{} achieves strong performance over baselines.} Notably, \ours{}-8B outperforms baselines with similar sizes by 9.51\%-18.72\% with tools and 2.03\%-11.24\% even \emph{without} tools, respectively.
\textbf{(ii) RL is critical for boosting TIR in VLMs.} Simply augmenting VLMs with tools without explicit reasoning supervision degrades accuracy. By contrast, RL delivers substantial gains, indicating that base checkpoints do not exhibit robust TIR ability and that RL is crucial for unlocking tool‑use capabilities in visual reasoning. Moreover, RL confers strong generalization: 
for example, \ours{}-8B attains accuracy on out‑of‑distribution VQA benchmarks comparable to substantially larger proprietary models (\eg, GPT‑o3 and Claude‑4.5‑Sonnet).
\textbf{(iii) \ours{} has a strong parameter efficiency.}
\ours{}-2B achieves competitive or even better performance with 8B baselines, while \ours{}-8B performs comparably to baselines with 38B parameters. 
The superiority of \ours{} demonstrates that \textbf{\env{} provides a scalable training ground for tool‑integrated agentic RL}, enabling robust visual reasoning in open‑source VLMs.

\begin{figure*}[t]
    \centering
    \begin{subfigure}[t]{0.33\linewidth}
        \centering
        \includegraphics[width=\linewidth]{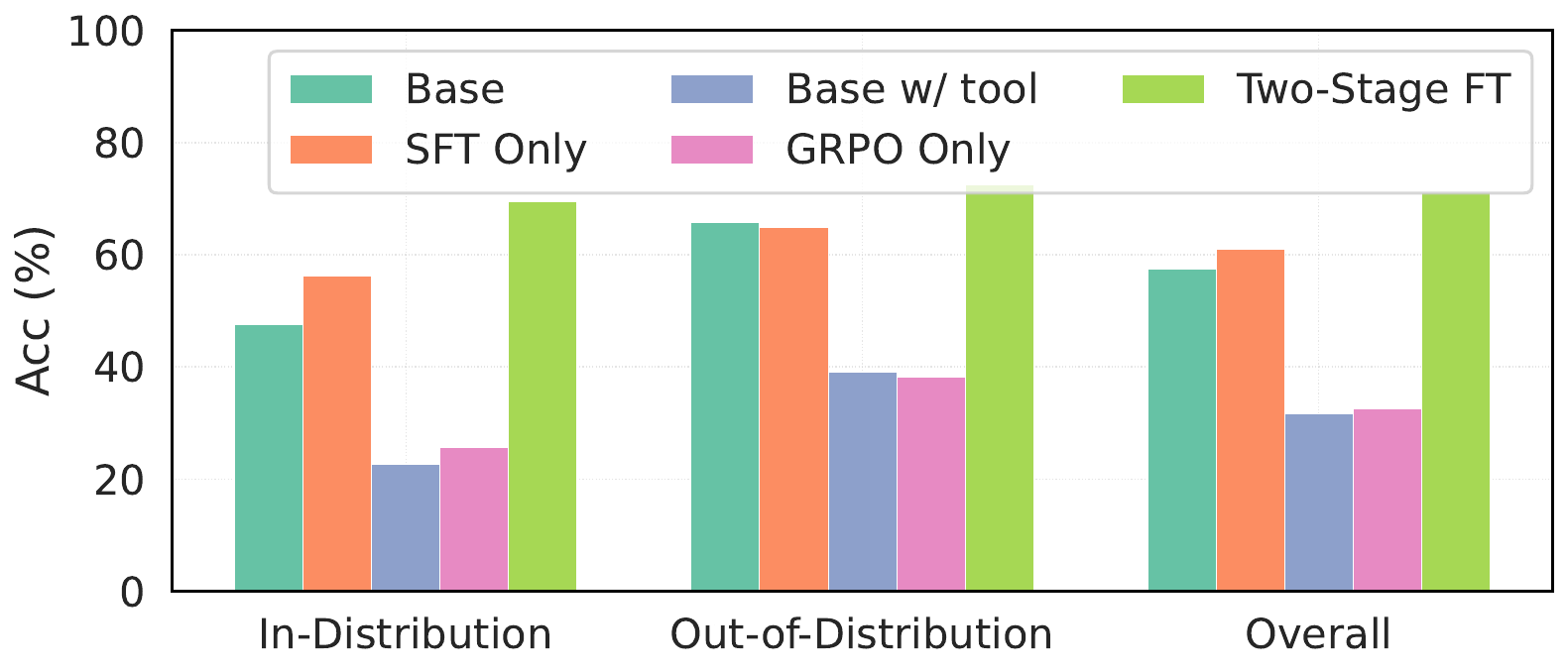}
        \caption{Effect of training stages}
        \label{fig:abl-training}
    \end{subfigure}
    \hfill
    \begin{subfigure}[t]{0.33\linewidth}
        \centering
        \includegraphics[width=\linewidth]{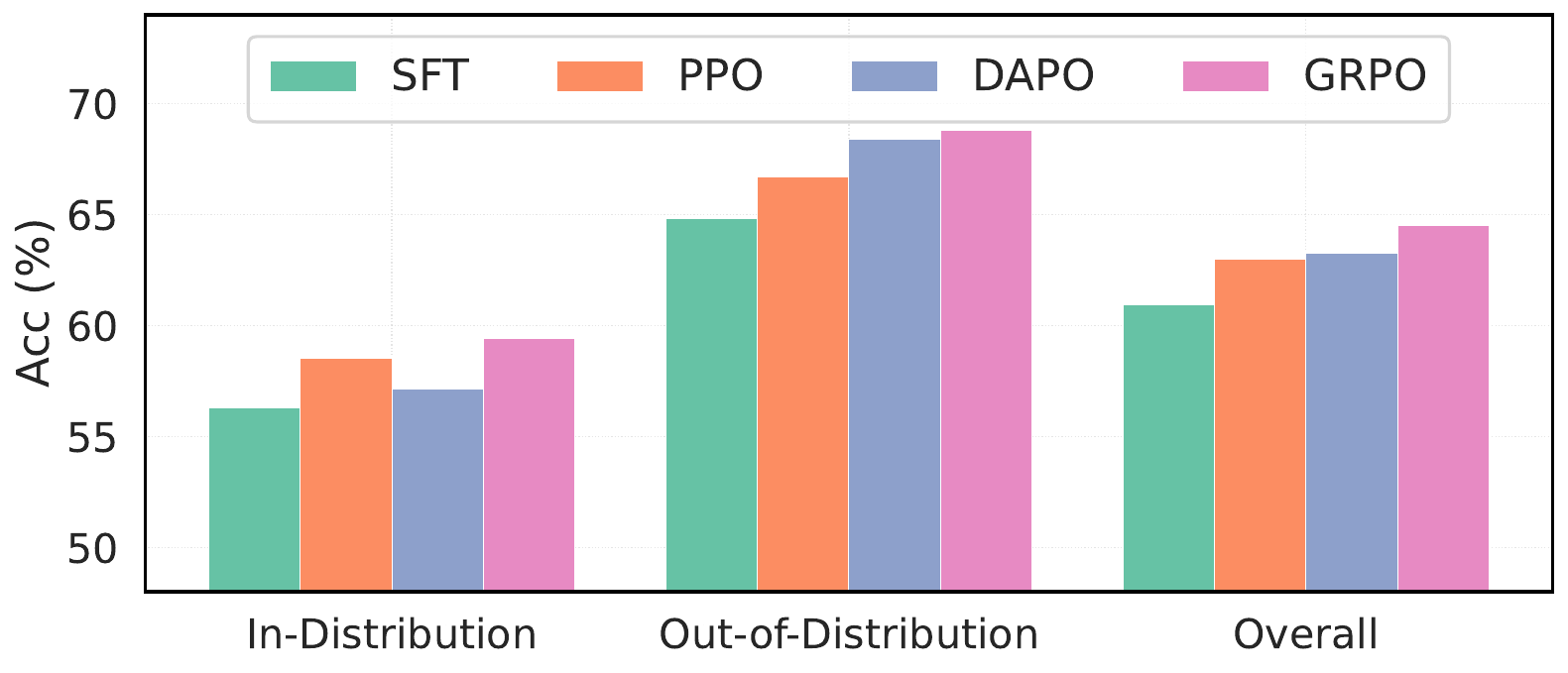}
        \caption{Effect of RL algorithms ($E$=100)}
        \label{fig:abl-rl}
    \end{subfigure}
    \hfill
    \begin{subfigure}[t]{0.33\linewidth}
        \centering
        \includegraphics[width=\linewidth]{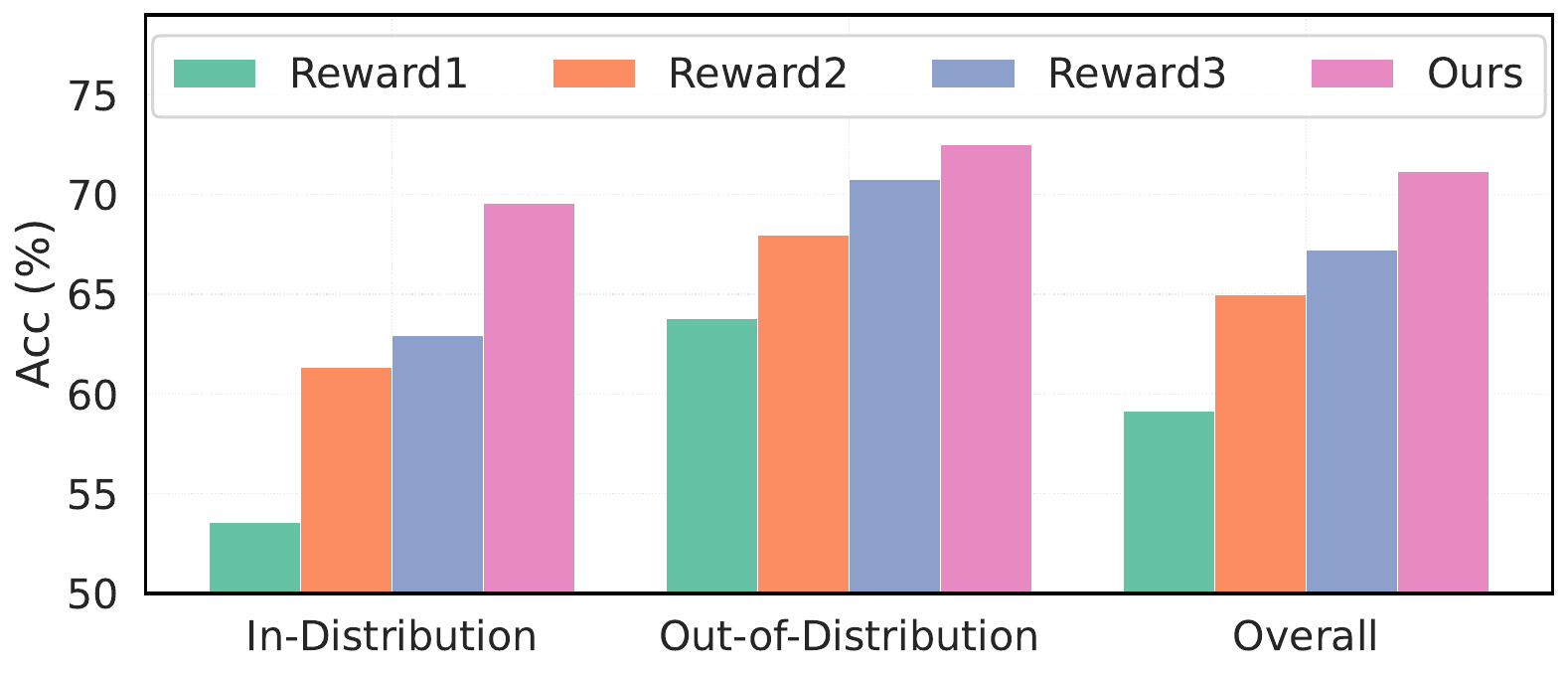}
        \caption{Effect of reward design}
        \label{fig:abl-reward}
    \end{subfigure}
       \hfill
        \begin{subfigure}[t]{0.33\linewidth}
        \centering
        \includegraphics[width=\linewidth]{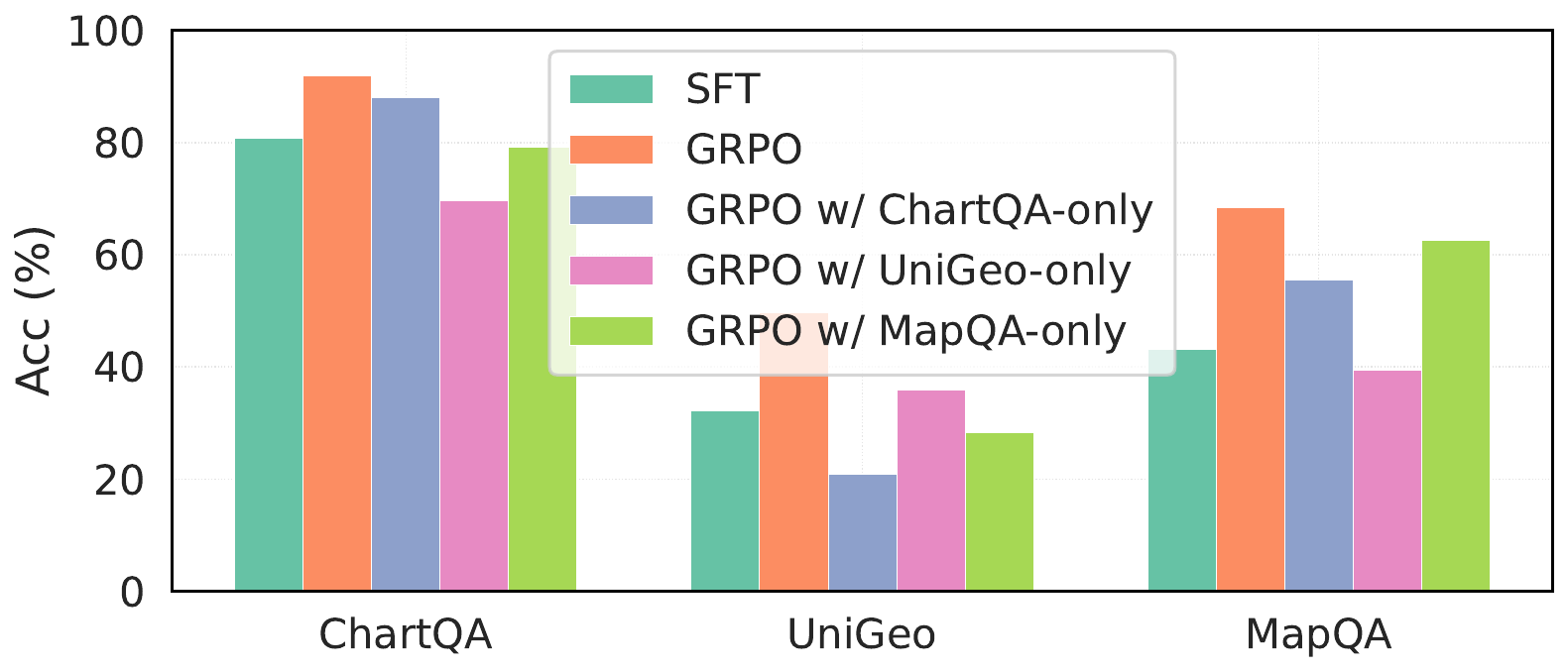}
        \caption{Effect of data diversity}
        \label{fig:adata-diversity}
    \end{subfigure}
    \hfill
    \begin{subfigure}[t]{0.33\linewidth}
        \centering
        \includegraphics[width=\linewidth]{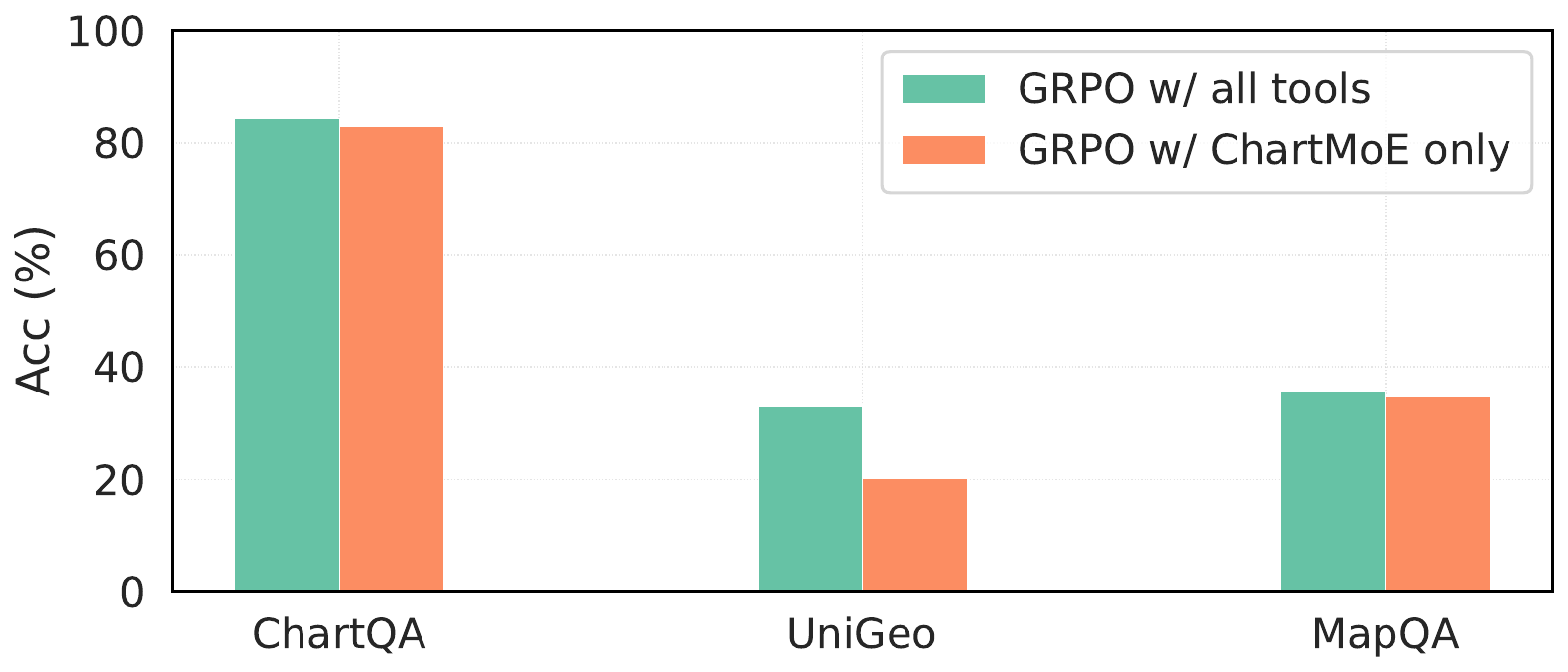}
        \caption{Effect of tool diversity ($E$=50)}
        \label{fig:tool-diversity}
    \end{subfigure}
    \hfill
    \begin{subfigure}[t]{0.33\linewidth}
        \centering
        \includegraphics[width=\linewidth]{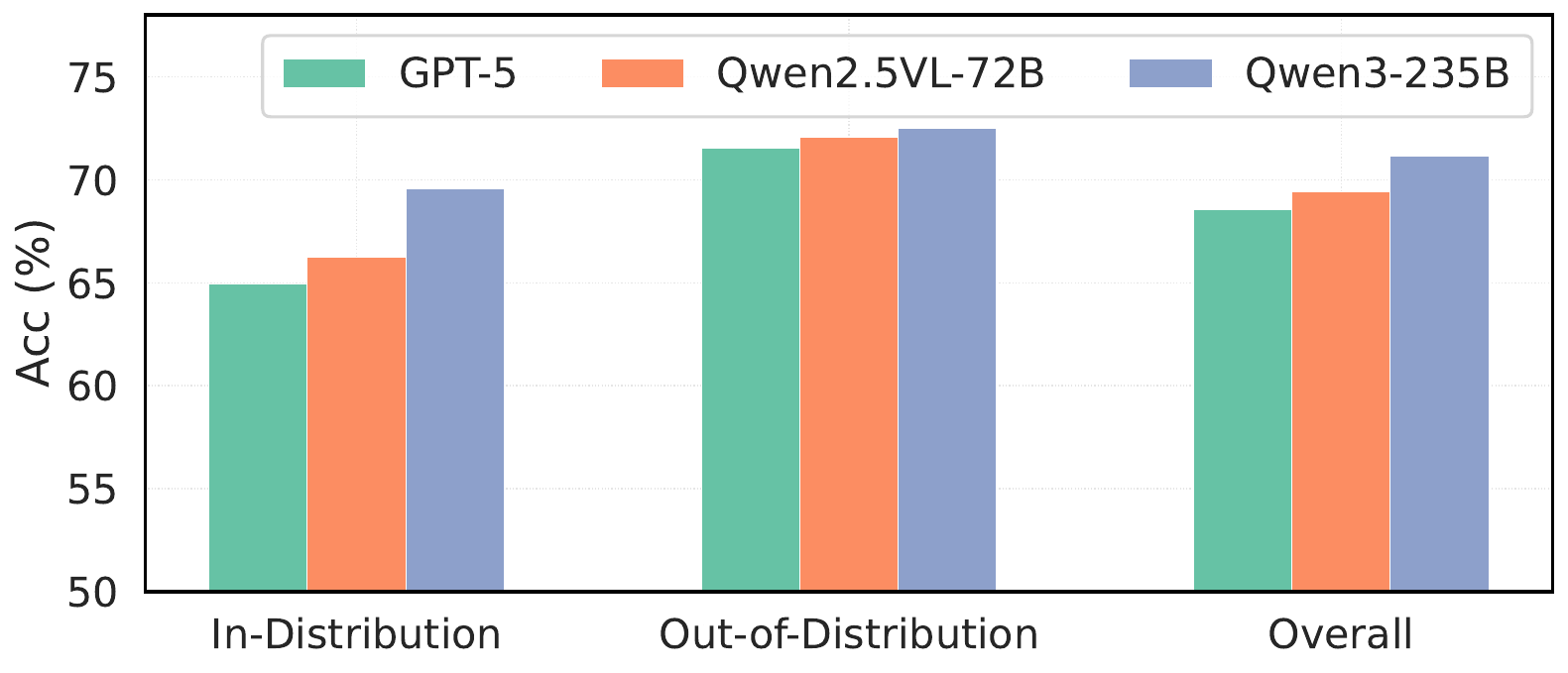}
        \caption{Effect of thinking trajectory quality}
        \label{fig:effect-of-traj-quality}
    \end{subfigure}
    \caption{Ablation studies and diversity analysis with InternVL3-8B as backbone VLM.}
    \label{fig:ablation-study}
\end{figure*}

\input{table/table-tir-fig1}

\subsection{Ablation Studies of \ours{}}
\noindent \textbf{Effect of Tool-Integrated Reasoning.}
Table~\ref{tab:main_results} also disentangles the roles of \emph{reasoning} and \emph{tool use} by reporting three variants: 
(1) \emph{w/o Tools} (63.66\%) that enables reasoning but disables tool access; 
(2) \emph{w/o Reasoning} (48.40\%) that exposes tools (with prior tool selection knowledge) without reinforcing reasoning; 
(3) and the \emph{full} \ours{} (71.14\%) that interleaves both. 
While strengthening reasoning or tool-use alone sometimes yields measurable gains, it remains suboptimal compared to \ours{}, which synergistically couples reasoning with tool execution.
Two key findings follow:
(i) simply exposing tools can be \emph{detrimental without sufficient reasoning} to navigate the action space, and
(ii) optimal problem solving arises when reasoning \emph{guides} (\emph{if/when/which}) tool invocation in an interleaved loop, rather than relying on either capability in isolation.

\noindent \textbf{Effect of Training Stage.}
Figure~\ref{fig:abl-training} shows that SFT serves as a critical \emph{warm-up}, aligning the model with tool-use formats and syntax (+3.46\%); RL then unlocks substantially larger gains (+10.19\% over SFT).
\emph{Both stages contribute}: SFT establishes the necessary behavioral priors for reliable tool interaction, while RL triggers deeper, interleaved reasoning required to navigate complex tasks effectively.

\noindent \textbf{Effect of RL Algorithm.}
In addition to GRPO, we evaluate PPO~\cite{schulman2017proximal} and DAPO~\cite{yu2025dapo} as preference-optimization algorithms; yet we observe that GRPO delivers the most robust performance, as shown in Figure~\ref{fig:abl-rl}.
The DAPO deficit may be due to the elimination of \emph{uniform‐reward} groups in its preference objective: in early stage, removing all-incorrect collapses the effective batch and starves the policy of learning signal; in later stage, removing all-correct erases useful supervision for reasoning–tool coordination. 
In contrast, GRPO retains all rollouts and uses group‑normalized advantages to deliver low‑variance, difficulty‑adaptive credit assignment, yielding more robust gains

\noindent \textbf{Effect of Reward Design.}
Figure~\ref{fig:abl-reward} compares different types of reward design, including (1) dense reward $R_{\text{dense}}=-1+0.5R_{\text{format}}+0.5R_{\text{correct}}+R_{\text{format}}\cdot R_{\text{correct}}$, (2) sparse reward $R_{\text{sparse}}=R_{\text{format}}\cdot R_{\text{correct}}$, (3) difficulty-adaptive reward $R_{\text{diff}}=R\cdot w$, where $w=\text{clamp}(2-D,0,1)\times0.5+0.5$ and $D$ represents the difficulty measured by the mean base reward across all rollouts, $D=\mathbb{E}[R]$, and (4) ours.
In comparison, our reward design couples a \emph{verifiable, low‑noise binary end signal with difficulty‑adaptive scaling} via group‑normalized advantages, yielding superior accuracy while avoiding overfitting to easy cases or exploitation of intermediate signals. 

\noindent \textbf{Effect of Tool and Reasoning in Vanilla VLMs.}
Table~\ref{tab:fig-tir} summarizes the effect of tools and reasoning on vanilla VLMs for open- and closed-source models. Directly augmenting VLMs with tools significantly degrades accuracy, and intrinsic chain-of-thought alone yields only modest gains on complex VQA. Supplying \emph{tool-selection priors} improves performance; however, gains are strongly task-dependent for commercial VLMs, and smaller open-source VLMs remain particularly challenged. These observations underscore that it is both important and non-trivial to endow VLMs with robust, generalizable tool-use capabilities that adapt across diverse visual reasoning scenarios. 

\begin{figure*}[t]
    \centering
    \begin{minipage}[t]{0.72\linewidth}
        \centering
        \includegraphics[width=\linewidth]{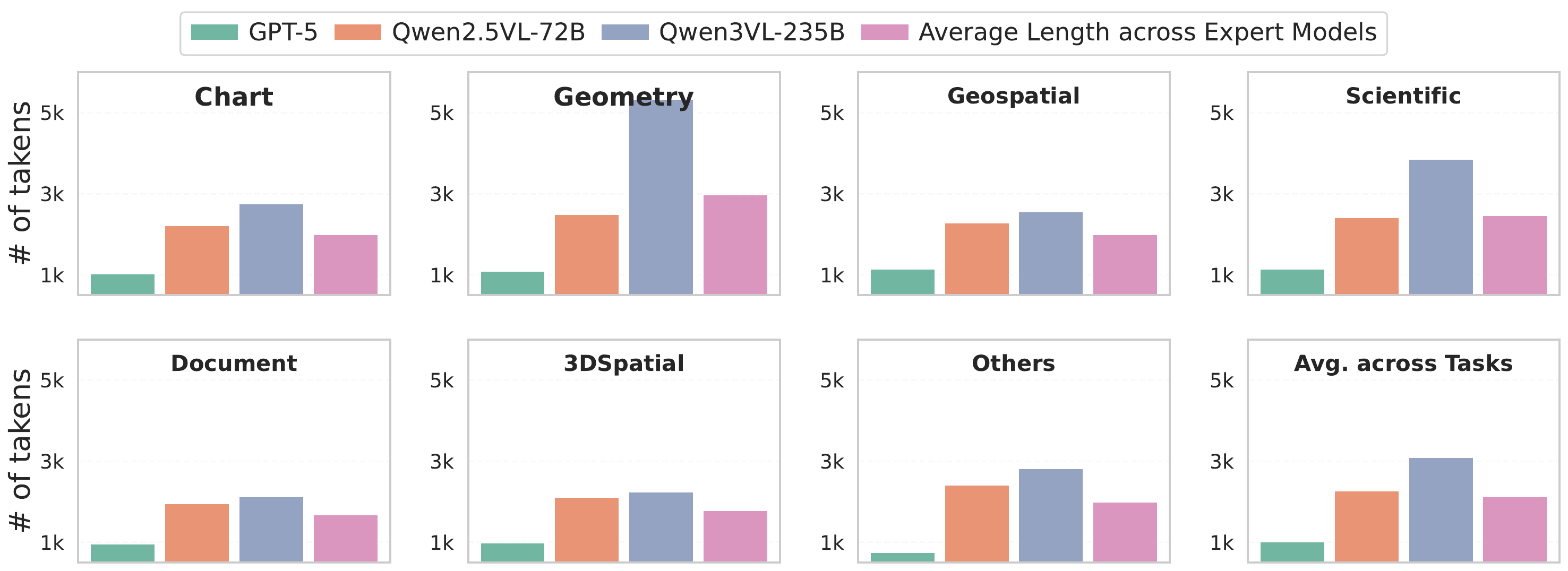}
        \caption{Quality of expert thinking trajectories by length.}
        \label{fig:traj-quality}
    \end{minipage}
    \hfill
    \begin{minipage}[t]{0.27\linewidth}
        \centering
        \includegraphics[width=\linewidth]{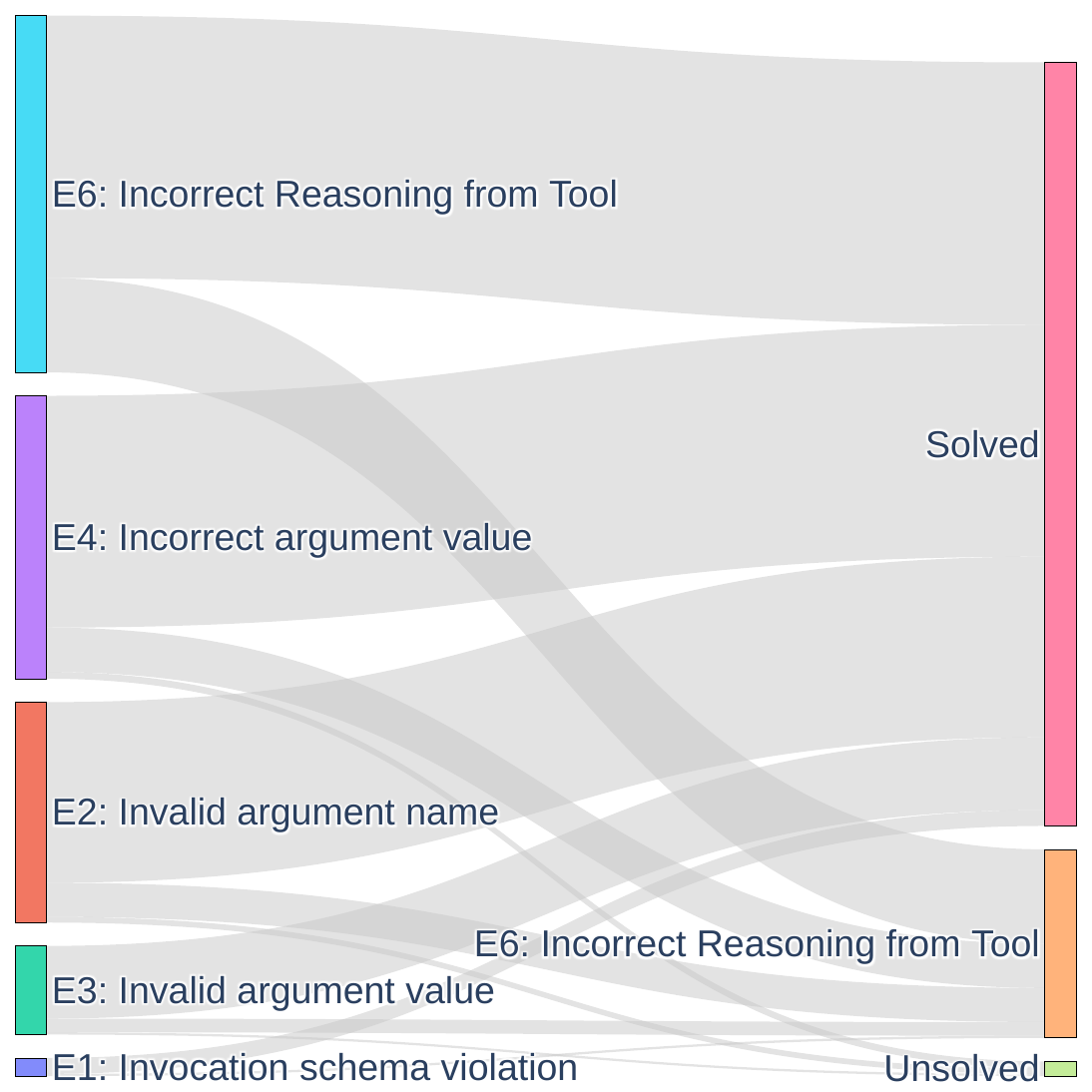}
        \caption{Error analysis.}
        \label{fig:error-transfer}
    \end{minipage}
\end{figure*}

\begin{figure}[t]
    \centering
    \begin{subfigure}[t]{0.49\linewidth}
        \centering
        \includegraphics[width=\linewidth]{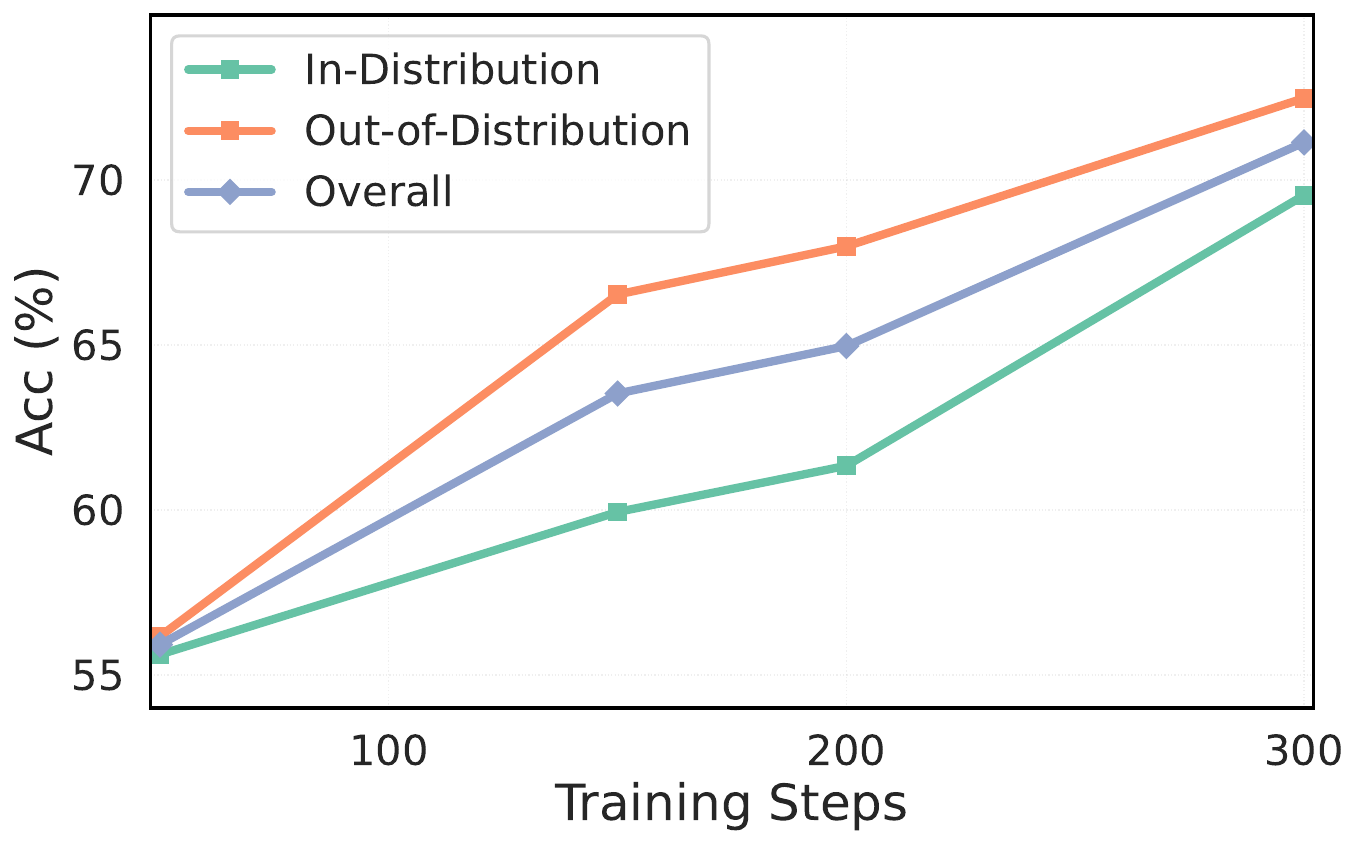}
        \caption{Training scaling}
        \label{fig:train-scale}
    \end{subfigure}
    \hfill
    \begin{subfigure}[t]{0.49\linewidth}
        \centering
    \includegraphics[width=\linewidth]{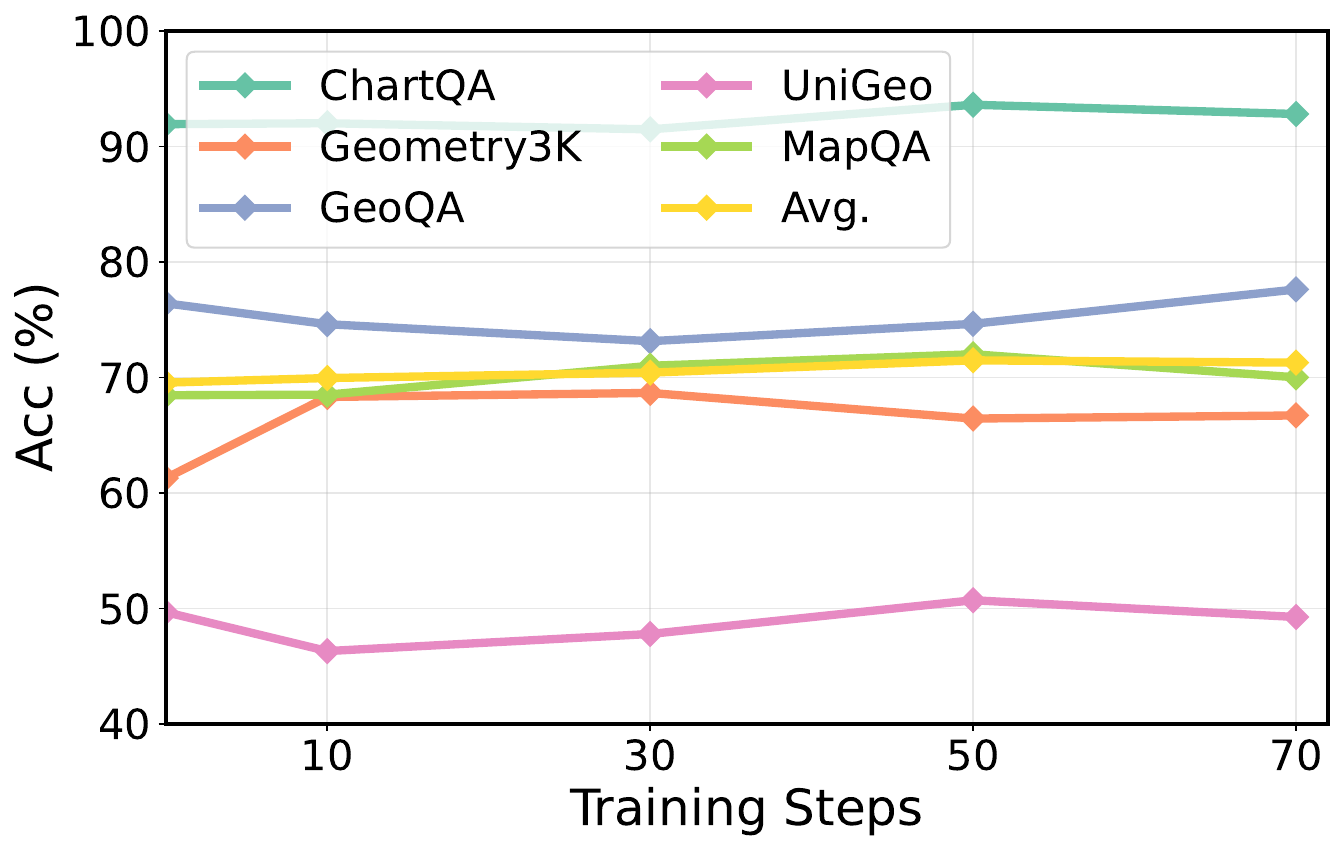}
    \caption{Tailpatch}
    \label{fig:tailpatch}
    \end{subfigure}
    \caption{Training scaling from easy to hard.}
    \label{fig:scale}
\end{figure}

\subsection{Additional Studies of \env{}}
\noindent \textbf{Data Diversity: Effect of Data Mixture.}
Figure~\ref{fig:adata-diversity} examines how task composition during RL shapes TIR. 
Single‑task training exhibits weak cross‑task transfer, especially across distinct domains such as ChartQA (chart understanding) and UniGeo (geometric reasoning), likely due to shifts in tool‑use schemas and argument distributions that narrow the learned policy’s coverage of the action space and impede general TIR. 
By contrast, multi-task training consistently improves generalization, indicating that exposure to a diverse task mixture during training is key to learning transferable TIR policies.

\noindent \textbf{Tool Diversity: Effect of Tool Mixture.}
Figure~\ref{fig:tool-diversity} analyzes how tool composition during RL shapes TIR using ChartMoE as an example. Training \emph{exclusively} with chart-understanding tools leaves chart-centric tasks (ChartQA; to a lesser extent, MapQA) largely unchanged, but transfers poorly to tasks with different tool affordances (UniGeo). Similar to data mixture, exposure to heterogeneous tool schemas and argument formats also mitigates over‑specialization and broadens the learned action policy.

\noindent \textbf{Reasoning Trajectory Quality.}
We assess the \emph{quality} of expert trajectories (GPT-5, Qwen2.5VL-72B, Qwen3VL-235B) using token length as a simple proxy in Figure~\ref{fig:traj-quality}. As expected, longer trajectories during the SFT warm‑up establish better behavior prior, thus contribute to higher performance after RL (Figure~\ref{fig:effect-of-traj-quality}). Note that commercial APIs such as GPT‑5 often do not expose full intermediate reasoning, yielding significantly shorter trajectories and weaker gains from agentic RL under the same training recipe.

\noindent \textbf{Tail-Patch Curriculum.}
To overcome the performance plateau, we adopt an easy-to-hard curriculum that \emph{tail-patch} training with increasingly challenging samples.
We use rollout pass rates from the previous checkpoint as a dynamic difficulty signal and isolate \emph{hard-but-learnable} instances with pass rates in $[0.125, 0.375]$.
Continuing RL exclusively on this high-difficulty slice pushes performance beyond the initial convergence point (69.54\% $\rightarrow$ 71.27\%) (Figure~\ref{fig:scale}).
This suggests a general principle: focusing the optimization budget on the \emph{frontier of competence}—cases where the model is uncertain yet occasionally succeeds—sharpens long-horizon reasoning and delivers fine-grained gains.

\subsection{Error Analysis and Case Study}
\noindent \textbf{Error Analysis.}
We re‑evaluate error patterns on the same 500 samples from Table~\ref{tab:error-def} after InternVL3-8B (left in Figure~\ref{fig:error-transfer}) training with \env{} (right). Agentic training within \env{} markedly resolves the majority of tool-call and reasoning failures observed in base models.

\begin{figure}[t]
    \centering
    \includegraphics[width=\linewidth]{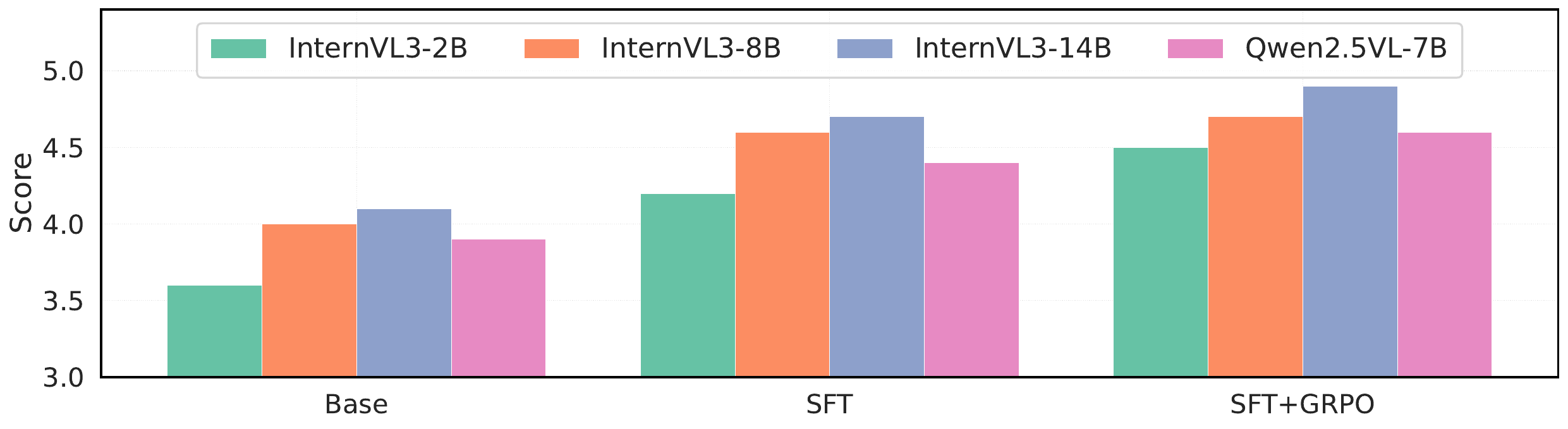}
    \caption{Human study on trajectory quality.}
    \label{fig:human-study}
\end{figure}

\noindent \textbf{Human Study for Trajectory Quality.}
As TIR trajectories are a key component of \env{}, we evaluate the quality of the generated interleaved tool-use and reasoning steps. Figure~\ref{fig:human-study} reports average human ratings (1–5 scale) over 40 randomly sampled solution trajectory per dataset. We observe that both training stages significantly improve TIR across tasks and model sizes.

\input{table/table-case}

\noindent \textbf{Case Studies.}
Table~\ref{tab:case_study} presents four case studies illustrating the capabilities acquired through agentic training in \env{}: (i) schema‑ and argument‑correct \emph{tool invocation}; (ii) \emph{post‑tool reasoning} that grounds the final answer in tool outputs; (iii) \emph{cross‑tool generalization} that repurposes atypical yet useful tools (\eg, applying ChartMoE on geometry problems); and (iv) \emph{robustness to imperfect tool outputs}, where the agent identifies and exploits partial signals over multiple turns to complete multi‑step reasoning.

%% file: table/table-main.tex
\begin{table*}[t]
\centering
\renewcommand\arraystretch{0.93}
\caption{Main results (acc\%) on five in-distribution and six out-of-distribution VQA benchmarks. $^\dagger$ indicates results reported from the original papers. 
\emph{w/o Tools} excludes tool access from both training and inference stage; \emph{w/o Reasoning} removes RL training stage.
}
\resizebox{1.0\linewidth}{!}{ %
\begin{tabular}{l @{\hskip2.5pt} | c c c c c | c | c c c c c c  | c | c c }
\toprule
\textbf{Baselines ($\downarrow$)} &  \multicolumn{6}{c|}{\textbf{In-Distribution}} & \multicolumn{7}{c|}{\textbf{Out-of-Distribution}} & \textbf{All} \\
\cmidrule(lr){2-7} \cmidrule(lr){8-14} \cmidrule(lr){15-15}
\textbf{Datasets ($\rightarrow)$} & ChartQA & Geometry3K & GeoQA & UniGeo & MapQA & avg. & TABMWP & AI2D & PlotQA & CLEVR-Math & IconQA & MathVista  &  avg. &  avg.  \\
\midrule
\multicolumn{10}{l}{\textit{Commercial VLMs (For reference)}} \\
\midrule
GPT-5 & 85.92 &	94.84 &	90.91 &	49.34 &	60.95 &	76.39 &	99.90 &	86.10 &	72.90 &	26.00 &	87.20 &	80.20 &	75.38 &	75.84 \\
GPT-5-mini & 82.80 &	91.01 &	88.01 &	46.02 &	61.35 &	73.84 &	98.57 &	85.85 &	70.95 &	27.15 &	87.73 & 79.30 &	74.93 &	74.43\\
GPT-o4-mini & 85.24 &	90.02 &	89.17 &	40.72 &	62.25 &	73.48 &	96.73 &	84.38 &	73.35 &	27.35 &	86.53 &	78.00 &	74.39 &	73.98 \\ 
GPT-o3 & 85.63 &	92.35 &	90.14 &	50.27 &	54.25 &	74.53 &	99.18 &	79.58 &	71.85 &	25.55 &	81.87 &	75.40 &	72.24 &	73.28 \\ 
Gemini-2.5-Pro & 83.64 & 94.34 & 86.85 & 60.34 & 64.20 &	77.87 &	99.90 &	90.28 &	77.40 &	25.70 &	96.93 & 82.00 &	78.70 &	78.33 \\
Gemini-2.5-Flash & 85.85 &	92.85 &	88.78 &	46.15 &	53.35 &	73.40 &	97.96 &	84.50 &	68.75 &	23.45 &	84.73 &	77.90 &	72.88 &	73.12 \\
Claude-4.5-Sonnet & 85.03 &	85.19 &	92.26 &	55.57 &	91.85 &	81.98 &	99.49 &	80.44 &	82.95 &	23.95 &	94.50 &	75.10 &	76.07	& 78.76 \\
\midrule
\multicolumn{10}{l}{\textit{Base Size: < 7B parameters}} \\
\midrule
InternVL3-2B   & 24.08  &	40.26  &	40.21  &	21.88  &	16.35  &	28.56  &	82.41  &	63.59  &	38.94  &	17.90  &	60.13  &	32.70  &	49.28  &	39.86 \\
\rowcolor{magenta!12} \ours{} (InternVL3-2B) &  88.55 &	56.24 &	68.54 &	42.31 &	33.35 &	57.80 &	85.35 &	73.80 &	63.85 &	39.81 &	84.13 &	60.00 &	67.82 &	63.27 \\
\rowcolor{magenta!3} \quad w/o Tools  &	72.88  &	43.93  &	66.30  &	39.52  &	39.45  &	52.42  &	74.23  &	64.70  &	35.90  &	15.93  &	63.53  &	49.60  &	50.65  &	51.45 \\
\rowcolor{magenta!3} \quad w/o Reasoning & 43.76	& 43.43	& 54.93	& 23.47 &	25.05 &	38.13 &	85.27 &	51.29 &	30.85 &	21.24 &	68.53 &	28.30 & 47.58 &	43.28 	\\
Qwen2.5-VL-3B  & 40.08  &	38.09  &	34.40  &	19.00  &	18.00  &	29.91  &	82.05  &	64.69  &	40.02  &	23.07  &	41.00  &	34.60  &	47.57  &	39.51 \\
LLaVA‑OneVision-1.5-4B 	& 61.60	&	50.92	&	29.79	&	14.06	&	27.25	&	36.72	&	93.15	&	73.68	&	43.90	&	24.00	&	84.80	&	44.50	&	60.67	&	49.79\\
\midrule
\multicolumn{10}{l}{\textit{Large Size: 7 - 13B parameters}}  \\
\midrule
Qwen2.5-VL-7B & 76.40 &	39.43 &	41.39 &	30.64 &	26.30 &	42.83 &	80.64 &	70.44 &	65.53 &	25.18 &	61.45 &	59.20 &	60.41 &	52.42 \\
\rowcolor{teal!15}  \ours{} (Qwen2.5-VL-7B) & 90.08 &	53.27 & 71.32 &	44.43 &	66.85 &	65.19 &	86.40 &	72.68 &	74.82 &	33.37 &	72.51 &	63.00 &	67.13 &	66.25  \\ 
\rowcolor{teal!4}  \quad w/o Tools & 83.32 &	48.92 &	68.11 &	33.42 &	51.35 &	57.02 &	64.13 &	56.09 &	76.45 &	26.92 &	68.58 &	56.10 &	58.05 &	57.58 \\	
\rowcolor{teal!4}  \quad w/o Reasoning & 79.68 &	47.92 &	58.61 &	34.88 &	26.70 &	49.56 &	82.11 &	72.09 &	66.90 &	17.93 &	82.87 &	42.50 &	60.73 &	55.65 \\					
VTool-R1-7B & 80.70$^\dagger$ & 62.06 &	65.18 &	33.29 &	19.65 &	52.18 &	64.21 &	79.34 &	48.85 &	19.70 &	75.67 &	36.60 &	54.06 &	53.20 \\
R1‑VL‑7B & 83.90$^\dagger$ &	58.90 &	62.09 &	37.93 &	23.87 &	57.34 &	92.64 &	76.51 &	48.50 &	18.00 &	87.27 &	63.50$^\dagger$ & 65.20 & 61.63 \\
R1‑Onevision‑7B  &59.92 &	55.79 &	57.47 &	28.30 &	33.20 &	46.94 &	84.19 &	61.75 &	35.40 &	19.30 &	79.30 &	64.10 &	57.34 &	52.61 \\	
Perception‑R1-7B & 	76.41 &	48.59 &	51.23 &	21.22 &	40.52 &	47.59 &	84.05 &	67.77 &	44.30 &	18.00 &	81.87 & 74.20$^\dagger$ &	61.70	& 55.29 \\	
InternVL3-8B   & 77.32 & 47.09 &	44.87 &	33.71 &	34.70 &	47.54 &	95.32 &	74.54 &	63.05 &	28.95 &	73.27 &	59.60 &	65.79 &	57.49 \\ 
\rowcolor{teal!15}  \ours{} (InternVL3-8B) &  91.92 &	61.27 &	76.41 &	49.67 &	68.45 &	69.54 &	96.52 &	79.86 &	66.38 &	38.20 &	91.09 &	62.80 &	72.48 &	71.14 \\ 
\rowcolor{teal!4}  \quad w/o Tool  & 84.24 &	51.47 &	73.10 &	34.70 &	53.50 &	59.40 &	94.68 &	75.37 &	60.25 &	29.24 &	80.93 &	62.80 &	67.21 &	63.66 \\
\rowcolor{teal!4}  \quad w/o Reasoning  & 68.56 & 35.64 &	55.90 &	27.06 &	26.45 &	42.72 &	92.34 &	50.45 &	38.70 &	28.18 &	80.07 &	29.10 &	53.14 &	48.40 \\ 
LLaVA‑OneVision-1.5-8B & 64.96 & 54.08 & 32.69 & 16.05 & 32.40 & 40.04 &	94.27 &	77.37 &	46.60 &	25.00 &	89.67 &	47.70 &	63.44 &	52.80 \\
\midrule
\multicolumn{10}{l}{\textit{XL Size: > 13B parameters}}  \\
\midrule
InternVL3-14B &  87.30 &	68.89 &	75.05 &	41.78 &	50.77 &	64.76 &	97.71 &	75.40 &	74.12 &	32.83 &	81.47 &	71.67 &	72.20 &	68.82 \\
\rowcolor{blue!12}  \ours{} (InternVL3-14B) & 93.60 &	83.87 &	78.85 &	52.94 &	70.50 &	75.95 &	98.02 &	87.60 &	76.00 &	40.86 &	92.13 &	68.00 &	77.10 &	76.58   \\
Qwen2.5-VL-32B  & 87.30&	65.22&	50.29&	42.94&	89.20&	66.99&	94.79&	82.16&	71.17&	35.14&	92.60&	68.20&	74.01&	70.82	 \\
InternVL3-38B &  89.20 &	70.72 &	79.11 &	45.14 &	55.13 &	67.86 &	98.35 &	83.03 &	78.48 &	33.88 &	83.73 &	74.03 &	75.25 &	71.89 \\
InternVL3-78B  & 89.70&	78.76&	85.43&	53.40&	58.15&	73.09&	99.39&	85.85&	82.35&	25.15&	86.33&	78.80&	76.31& 74.85 \\


\bottomrule
\end{tabular}%
}
\label{tab:main_results}
\end{table*}

%% file: table/table-tir-fig1.tex
\begin{table*}[t]
\centering
\renewcommand\arraystretch{0.93}
\caption{Effect of reasoning and tool-use results without RL training. \emph{w/ Tools} refers to directly augmenting VLMs with tools significantly degrades accuracy; \emph{w/ Reasoning} refers to CoT reasoning without tools. \emph{w/ T\&R} refers to a different setting compared to TIR, only supplying \emph{tool‑selection prior knowledge} and interleaving reasoning with tool execution improve performance. }
\resizebox{1.0\linewidth}{!}{ %
\begin{tabular}{l @{\hskip2.5pt} | c c c c c | c | c c c c c c  | c | c c }
\toprule
\textbf{Baselines ($\downarrow$)} &  \multicolumn{6}{c|}{\textbf{In-Distribution}} & \multicolumn{7}{c|}{\textbf{Out-of-Distribution}} & \textbf{All} \\
\cmidrule(lr){2-7} \cmidrule(lr){8-14} \cmidrule(lr){15-15}
\textbf{Datasets ($\rightarrow)$} & ChartQA & Geometry3K & GeoQA & UniGeo & MapQA & avg. & TABMWP & AI2D & PlotQA & CLEVR-Math & IconQA & MathVista  &  avg. &  avg.  \\
\midrule
\rowcolor{gray!15}  GPT-5 & 85.92 &	94.84 &	90.91 &	49.34 &	60.95 &	76.39 &	99.90 &	86.10 &	72.90 &	26.00 &	87.20 &	80.20 &	75.38 &	75.84 \\
\quad w/ Tools & 58.60 & 74.04 & 65.00 & 28.78& 31.20 & 51.52 &69.83 & 65.07 & 46.75 & 18.96 & 60.53 & 42.60 & 50.62 & 51.03\\
\quad w/ Reasoning  & 85.68 & 91.01 & 86.27 & 51.25& 59.48 & 74.74 &99.81 & 83.03 & 84.73& 25.50 & 86.27 & 80.90 & 76.71 & 75.81\\
\quad w/ T\&R &  94.00 & 93.84 & 91.56 & 51.69 & 57.35& 77.69 & 98.92 & 84.62& 79.45 & 27.79& 91.07 & 57.80 & 73.28 & 75.28 \\
\midrule
\rowcolor{gray!15}  InternVL3-8B   & 77.32 & 47.09 &	44.87 &	33.71 &	34.70 &	47.54 &	95.32 &	74.54 &	63.05 &	28.95 &	73.27 &	59.60 &	65.79 &	57.49 \\ 
\quad w/ Tools & 25.44 &  17.06& 27.27 & 26.90& 16.22 & 22.58 &58.15 & 51.74 & 58.31 & 18.01 & 25.86 &  22.70& 39.13 & 31.61\\
\quad w/ Reasoning  & 81.28& 47.09 & 45.07 & 35.81&31.10 &  48.07& 94.38&76.51 & 65.85& 24.15 &75.53  &60.70  & 66.19 &57.95\\
\quad w/ T\&R & 68.56 & 35.64 & 55.90& 27.06 & 26.45&42.72 & 92.34 & 50.45& 38.70& 28.18& 80.07& 29.10& 53.14 &48.40 \\
\bottomrule
\end{tabular}%
}
\label{tab:fig-tir}
\end{table*}

%% file: table/table-case.tex
\begin{table*}[t]
\centering
\renewcommand\arraystretch{0.95}
\caption{Case studies on geometric reasoning tasks with \ours-8B trained in \env{}.}
\resizebox{\linewidth}{!}{
    \begin{tabular}{p{1.8cm}p{21cm}}
       \toprule
       \multicolumn{2}{l}{\textbf{Case Study 1}: Find m $\angle$ R. \textbf{Answer: A}: 58} \\
        \midrule
        \bf \multirow{6}{*}{\includegraphics[width=\linewidth]{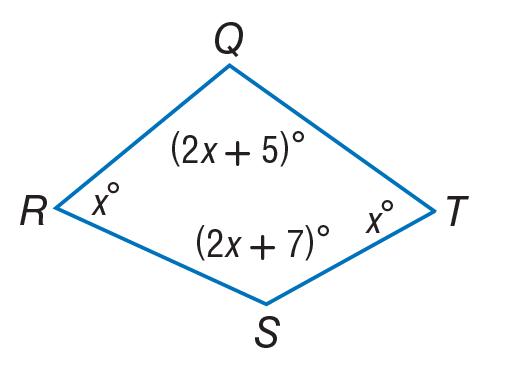}} & \textbf{InternVL3-8B-Tool}: \textcolor{blue}{<tool\_call>}\{"tool": "multimath", "task": "solve\_the\_problem", "arguments": \{"image\_url": <image\_url>, "data": \{\textcolor{red}{"image/png": base64, iVBORw0KGgoAAAANSUhEUgAAAOAAAAHCAIAAADYXJpAAAAA3NCSV $\cdots$ (including long base64 image data)}\}\}\textcolor{blue}{</tool\_call>} \hfill \colorbox{red!15}{\textbf{Tool Call Error}: \emph{Invalid Arguments}. (\ding{56})} \\
        & \textbf{\ours}: \textcolor{blue}{<tool\_call>}\{"tool": "gllava", "task": "solve", "arguments": "Find m $\angle R$ from the diagram: a diamond-shaped quadrilateral with vertices R (left), Q (top), T (right), S (bottom). Angles labeled: at $R = x^\circ$, at $T = x^\circ$, at $Q = (2x+5)^\circ$, at $S = (2x+7)^\circ$."\}.\textcolor{blue}{</tool\_call>} \hfill \colorbox{green!15}{\textbf{Tool Call}: \emph{Correct} (\ding{51})} \\
        \midrule
        \multicolumn{2}{p{23cm}}{\textbf{Case Study 2}: As shown in the figure, in $\odot O$, $AB$ is a diameter and $CD$ is a chord. $AB \perp CD$, with the foot of the perpendicular at point $E$. Connect $CO$ and $AD$. If $\angle BOC = 30^\circ$, then the degree measure of $\angle BAD$ is ( ). \textbf{Answer: 15$^\circ$}} \\
        \midrule
        \bf \multirow{6}{*}{\includegraphics[width=\linewidth]{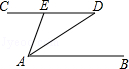}}  & \textbf{InternVL3-8B-Tool}: $\cdots$\textasciigrave\textasciigrave\textasciigrave output $\cdots$Step 12 (Property of isosceles triangle): $\angle BAD = \angle OAD$. Answer: $\angle BAD = 15^\circ$.\textasciigrave\textasciigrave\textasciigrave \textcolor{blue}{<think>} Alright, \textcolor{red}{let's tackle this geometry problem step by step}. $\cdots$ But according to the tool output, the answer is 15 degrees, which is option D. \textcolor{red}{But my computation also says the answer is 52.5 degrees. So the answer is closer to C}. \textcolor{blue}{</think>}$\cdots$ \hfill \colorbox{red!15}{\textbf{Correct Tool, Wrong Reasoning}: \emph{Incorrect reasoning to answer.}. (\ding{56})}\\
        & \textbf{\ours}: $\cdots$ \textasciigrave\textasciigrave\textasciigrave output $\cdots$Step 12 (Property of isosceles triangle): $\angle BAD = \angle OAD$. Answer: $\angle BAD = 15^\circ$.\textasciigrave\textasciigrave\textasciigrave\textcolor{blue}{<think>} \textcolor{green!50!black}{$\cdots$Based on the previous verification and tool output, I can analyze and provide the answer of $15^\circ$.}\textcolor{blue}{</think> <answer>}15$^\circ$\textcolor{blue}{</answer>}. \hfill \colorbox{green!15}{\textbf{Reasoning}: \emph{Correct} (\ding{51})} \\
        \midrule
        \multicolumn{2}{p{23cm}}{\textbf{Case Study 3}: $AB || CD$. Line $EF$ intersects $AB$ at point $E$ and intersects $CD$ at point $F$. $EG$ bisects $\angle BEF$ and intersects $CD$ at point $G$. If $\angle 1 = 50^\circ$, then $\angle 2$ equals ( ). Choices:A: $50^\circ$B: $60^\circ$C: $65^\circ$D: $90^\circ$. \textbf{Answer: C: 90$^\circ$.}} \\
        \midrule
        \bf \multirow{7}{*}{\includegraphics[width=\linewidth]{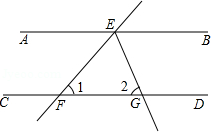}} & \textbf{\ours}: \textcolor{blue}{<think>} $\cdots$ \textcolor{green!50!black}{ChartMoE is appropriate because it is designed to analyze structured geometric diagrams, accurately reading angle labels, identifying parallel-line configurations, and aligning textual annotations with diagram elements. I will have it parse the diagram to confirm $AB||CD$, locate the transversal EF and the points E, F, G, and verify that EG bisects $\angle BEF$ while intersecting CD at G.} By interpreting the angles formed at E relative to BE and FE and using the parallel-line angle relations (corresponding, alternate interior, and consecutive interior angles), ChartMoE will relate $\angle 1$ to the angle between BE and EG, propagate the bisector relation to $\angle BEF$, and then deduce the measure of $\angle 2$ at the intersection of EF and CD from the resulting angle measures. Finally, it will align that computed value with the provided choices. Therefore, I will call chartmoe now.\textcolor{blue}{</think><tool\_call>}\{"tool": "chartmoe", "task": "analyze"\}\textcolor{blue}{</tool\_call>} $\cdots$ \hfill \colorbox{green!15}{\textbf{New Tool Selection}: \emph{Expected G-LLaVA but got ChartMoE.} (\ding{51})}\\
        \midrule
        \multicolumn{2}{p{23cm}}{\textbf{Case Study 4}: Find X Z. \textbf{Answer: D: 34.}} \\
        \midrule
        \bf \multirow{16}{*}{\includegraphics[width=\linewidth]{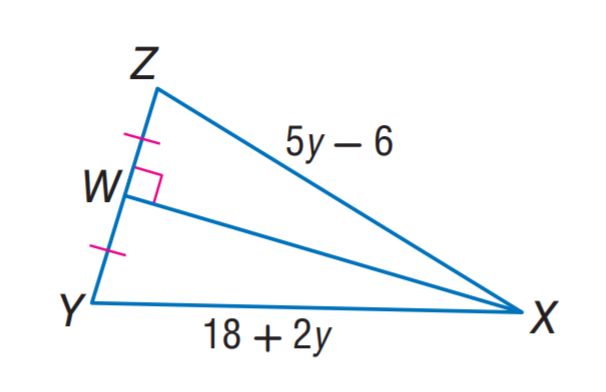}} & \textbf{InternVL3-8B-Tool}: \textcolor{blue}{<tool\_call>}\{"tool": "multimath", "task": "Solve the problem", "arguments": \{"image": "<image>", "question": "Find X Z. $\cdots$"\}\}\textcolor{blue}{</tool\_call>}\textasciigrave\textasciigrave\textasciigrave output $\cdots$ Step 5 (Simplify the equation): $XZ^2 = 29y^2 + 12y + 360.$\textbackslash nAnswer: \textbackslash boxed\{$29y^2 + 12y + 360$\}\textasciigrave\textasciigrave\textasciigrave\textcolor{blue}{<think>} $\cdots$ But as we saw earlier, this leads to a quadratic with no real solutions.\textbackslash n \textcolor{red}{Wait, maybe I made a mistake in the tool output. Let me double-check the steps: \textbackslash n Step 3: $XZ^2 = (5y - 6)^2 + (18 + 2y)^2 = 25y^2 - 60y + 36 + 324 + 72y + 4y^2 = 29y^2 + 12y + 360$. That seems correct.}\textbackslash n $\cdots$ Step 5: Simplify the equation: $XZ^2 = 29y^2 + 12$\textcolor{blue}{</think><answer>} \textbackslash boxed\{C\}.\textcolor{blue}{</answer>} \hfill \colorbox{red!15}{\textbf{Tool Output Error}: \emph{Incorrect reasoning to answer.}. (\ding{56})}\\
        & \textbf{InternVL3-8B-Reasoning}: <think> Alright, let's tackle this problem step by step. First, I notice that triangle WYZ is a right triangle because it has a right angle at W. $\cdots$ Using the Pythagorean theorem again:\textbackslash n$XZ^2 = WZ^2 + XY^2$\textbackslash n$XZ^2 = 34^2 + 34^2$, $XZ^2=2312$, $XZ=\sqrt{2312}=34*\sqrt{2}$\textbackslash n Therefore, the length of XZ is $34\sqrt{2}$.\textbackslash n</think>\textbackslash n<answer>$34\sqrt{2}$</answer>" \hfill \colorbox{red!15}{\textbf{No Tool-use}: \emph{Insufficient capability to answer.}. (\ding{56})}\\
        & \textbf{\ours}: \textcolor{blue}{<think>} $\cdots$ \textcolor{green!50!black}{Multimath is appropriate because} $\cdots$ I will read from the image that $WX\perp ZX$ $\cdots$. \textcolor{green!50!black}{Multimath will translate these visual relations into algebraic constraints}, $\cdots$. \textcolor{blue}{</think><tool\_call>}\{"tool": "multimath", "task": "solve"\}\textcolor{blue}{</tool\_call>}\textasciigrave\textasciigrave\textasciigrave output $\cdots$ Step 10 (Calculate the value of XZ): $XZ = 18 + 2y = 18 + 2(\frac{6}{5}) = 18 + \frac{12}{5} = \frac{90}{5} + \frac{12}{5} = \frac{102}{5} = 20.4$. \textasciigrave\textasciigrave\textasciigrave \textcolor{blue}{<think>} Looking at the tool's solution, it set up the equation $25y^2 - 60y + 36 = 0$ and solved to get $y = 6/5$. Let me verify this step by step. $\cdots$ \textcolor{green!50!black}{Wait, let me check if this makes sense geometrically.} $\cdots$ Actually, if the tick marks indicate equal segments, shouldn't we have $5y - 6 = 18 + 2y$? Let me solve this: $5y - 6 = 18 + 2y 3y = 24 y = 8$ With $y = 8$, let's calculate XZ. If $XZ = 18 + 2y$, then: $XZ = 18 + 2(8) = 18 + 16 = 34$. \textcolor{green!50!black}{Hmm, that's different from the tool's answer of $20.4$. Let me reconsider.} $\cdots$ \textcolor{green!50!black}{The tool calculated 20.4, but based on the equal segment marks, the answer is $34$.}\textcolor{blue}{</think><answer>}$34$\textcolor{blue}{</answer>}" \hfill \colorbox{green!15}{\textbf{TIR}: \emph{correct} (\ding{51})} \\
       \bottomrule
    \end{tabular}
}
	\label{tab:case_study}
\end{table*}

%% file: section/5-conclusion.tex
\section{Conclusion}
\label{sec:conclusion}
In this work, we introduce \env{}, a scalable training ground for tool‑integrated, visual‑centric agentic RL in VLMs, and \ours{}, an agent trained to interleave multi‑turn reasoning with structured tool use. 
Practitioners leveraging \env{} for boosting TIR in VLMs may consider: (i) train policies that \emph{interleave} reasoning with tools; (ii) warm up with SFT for tool call syntax/format, then apply online RL for in-depth tool-integrated visual reasoning; (iii) diversify tasks and tools to expand action‑space coverage; and (iv) allocate additional RL budget to hard‑but‑learnable slices via tail‑patching. 
Our verifiers emphasize terminal correctness and structural validity; richer stepwise semantics and broader tool ecosystems may further benefit long‑horizon TIR. 
\env{} provides a scalable agentic training environment—unified interfaces, executable feedback, and efficient logging—for systematic progress in \emph{thinking with images}. 

%% file: section/6-appendix.tex
\clearpage
\setcounter{page}{1}
\maketitlesupplementary

\section{Task and Data Information}
\label{app:data}
\input{appendix/app-data}

\section{Baseline Details}
\label{app:baseline}
\input{appendix/app-baseline}

\section{Prompt Templates}
\label{app:prompt}
\input{appendix/app-prompt}


\section{Additional Method Details of \env{}}
\label{app:method}
\noindent\textbf{Interface Details.}
\env{} follows the Gymnasium API with core methods \texttt{reset} and \texttt{step}. On \texttt{reset}, the environment samples a task \(x \in \mathcal{I}\) (question + image) and returns the initial observation. At time step \(t\), given the interaction history \(c_{t-1}\), the environment provides the current observation \(o_t\), and the agent produces a thought–action pair \((g_t, a_t)\). The environment executes \(a_t\), transitions according to the POMDP formalization in Section~\ref{sec:pre}, and yields the next observation \(o_{t+1}\), enabling multi-turn tool-integrated interaction.

\input{appendix/app-method}
\section{Additional Implementation Details}
\label{app:implementation}
\input{appendix/app-implmenetation}


%% file: appendix/app-data.tex
Table~\ref{tab:datastats} presents the statistics of the in‑distribution (ID) and out‑of‑distribution (OOD) datasets; the additional details are as follows:
\subsection{Information for In-Distribution Datasets}
We curate a diverse suite of training tasks integrated into \env, spanning fifteen datasets across eight distinct reasoning domains:
\begin{itemize}
    \item \textbf{Chart Understanding}: {FigureQA}~\cite{ebrahimi2018figureqa} and {ChartQA}~\cite{masry2022chartqa} demand precise quantitative extraction and logical inference over diverse data visualizations, requiring the agent to align visual features (bars, lines) with numerical values.
    \item \textbf{Geometric Reasoning}: {Geometry3K}~\cite{lu2021inter}, {GeoQA}~\cite{chen2021geoqa}, and {UniGeo}~\cite{chen2022unigeo} necessitate grounding symbolic theorem application in diagrammatic parsing to solve multi-step spatial and quantitative problems.
    \item \textbf{Mathematical Reasoning}: {MathVision}~\cite{wang2024measuringmultimodalmathematicalreasoning} and {MathV360}~\cite{shi2024mathllavabootstrappingmathematicalreasoning} evaluate the interpretation of complex mathematical diagrams and symbolic expressions, requiring the synthesis of visual perception with algebraic deduction.
    \item \textbf{Geospatial Reasoning}: {MapQA}~\cite{chang2022mapqa} and {InfographicVQA}~\cite{mathew2022infographicvqa} challenge agents to perform rigorous spatial lookups, legend-symbol alignment, and information retrieval across dense graphical layouts.
    \item \textbf{Visual Scientific Reasoning}: {ThinkVL}~\cite{deng2025openvlthinkercomplexvisionlanguagereasoning}, {VizWiz}~\cite{bigham2010vizwiz}, and {ScienceQA}~\cite{lu2022learn} encompass a broad spectrum of tasks ranging from accessibility-focused visual description to domain-specific multimodal scientific inquiry.
    \item \textbf{Document Understanding}: {DocVQA}~\cite{mathew2021docvqa} tests layout-aware information extraction from unstructured documents, including financial reports, forms, and scanned articles.
    \item \textbf{Spatial Reasoning}: {CLEVR}~\cite{johnson2017clevr} provides a diagnostic environment for compositional logic, assessing the agent's ability to execute complex reasoning chains regarding attribute recognition, spatial relations, and counting in 3D scenes.
    \item \textbf{General Visual Reasoning (Others)}: {A-OKVQA}~\cite{schwenk2022okvqa} requires the retrieval of external world knowledge and commonsense reasoning to address open-ended visual questions.
\end{itemize}

\subsection{Information for Out-of-Distribution Datasets}
To assess generalization, we evaluate on six additional benchmarks featuring distinct reasoning:
\begin{itemize}
    \item \textbf{TABMWP}: TABMWP~\cite{lu2022dynamic} focuses on table-based math word problems requiring cell selection, row/column aggregation, and arithmetic reasoning grounded in semi-structured text, testing symbolic computation beyond visual chart parsing.
    \item \textbf{AI2D}: AI2D~\cite{kembhavi2016diagram} is a diagram-centric science QA benchmark involving annotated figures (labels, arrows, parts); it stresses the joint interpretation of schematic layouts and textual cues via multi-hop reasoning.
    \item \textbf{PlotQA}: PlotQA~\cite{methani2020plotqa} centers on scientific plots (axes, legends, bars/lines) demanding high-precision value reading and aggregation, serving as a complementary but distributionally distinct test bed to ChartQA.
    \item \textbf{CLEVR-Math}: CLEVR-Math~\cite{lindstrom2022clevr} extends compositional reasoning in rendered 3D scenes with arithmetic operations, assessing program-like visual logic coupled with numeric computation.
    \item \textbf{IconQA}: IconQA~\cite{lu2021iconqa} comprises textbook-style diagram and icon questions (often multiple-choice) that emphasize abstract visual relations and symbolic understanding, diverging from natural image statistics.
    \item \textbf{MathVista}: MathVista~\cite{lu2023mathvista} is a holistic mathematical reasoning suite spanning real images, charts, and diagrams; it integrates OCR, quantitative reading, geometry, and multi-step deduction for comprehensive evaluation.
\end{itemize}

\input{table/tab-data}

\section{Toolset Information}
\label{app:tool}

  
In \env, we enable access to 26 tools from four categories detailed as follows: 

\noindent \textbf{Perception.}
The perception layer provides low‑level visual signals that supply structural cues for higher‑level reasoning, including:
\begin{itemize}
    \item \textbf{Detection}: \texttt{GroundingDINO} is an open-set object detection model that unifies visual and textual modalities, allowing for the detection of arbitrary objects described by natural language. It performs open‑set, text‑conditioned object detection, localizing arbitrary entities referenced by natural‑language queries.
    \item \textbf{Segmentation}: \texttt{SAM} (Segment Anything Model) is a foundational model for promptable image segmentation that can zero-shot segment any object based on flexible inputs like points, boxes, or text. It offers promptable, zero‑shot segmentation, isolating regions given points, boxes, or textual prompts.
    \item \textbf{OCR}: \texttt{EasyOCR} is a versatile and ready-to-use optical character recognition tool supporting over 80 languages and various writing scripts for robust text extraction from images. It conducts multilingual optical character recognition, extracting text from images across diverse scripts and layouts.
\end{itemize}

\noindent \textbf{Chart Understanding.}
The chart layer specializes in plot‑ and table‑centric inference. Chart understanding tool collections like \emph{ChartMoE} use a Mixture‑of‑Experts connector to enhance visual–text alignment, supporting both structured data extraction and high‑level analytical reasoning for chart question answering, including:
\begin{itemize}
    \item \textbf{Chart To Table.}
The \emph{ChartToTable} tool subset converts chart images into structured tabular data suitable for downstream numerical reasoning. 
\texttt{UniChart} and \texttt{DePlot} recover underlying data points by detecting axes, legends, and visual marks, while \texttt{ChartMoE.to\_table} and \texttt{ChartMoE.extract\_data} leverage the ChartMoE backbone to extract series-wise values and metadata (\eg, categories, units) with improved robustness across chart styles.

\item \textbf{Chart To SVG.}
The subset of \emph{ChartToSVG} tools reconstructs vectorized representations of charts. 
\texttt{OpenCV} provides low-level image processing to segment graphical primitives, and \texttt{ChartDet} localizes key chart components (axes, legends, bars, lines, markers); \texttt{ChartOCR} recognizes textual elements (titles, axis labels, tick labels, legends) and anchors them to detected regions, enabling SVG-style reconstruction.

\item \textbf{Chart To SCG.}
The subset of \emph{ChartToSCG} tools maps charts into structured scene graphs (SCG). 
\texttt{ChartDet} and \texttt{ChartOCR} are combined to instantiate nodes for visual and textual elements and edges for relations such as series membership, axis association, and legend–color bindings, yielding a graph representation amenable to symbolic reasoning.

    \item \textbf{Captioning Color.}
The subset of \emph{CaptioningColor} generates natural-language descriptions of charts with explicit grounding in visual encodings. 
\texttt{ChartAssistant} produces concise summaries of chart type, trends, and salient extrema, whereas \texttt{ChartVLM} provides more detailed captions that explicitly reference colors, legends, and value ranges.

    \item \textbf{QA Analysis.}
The \emph{QAAnalysis} supports chart question answering and intermediate analysis. 
\texttt{ChartMoE.answer} directly predicts answers to chart-related questions, while \texttt{ChartMoE.analyze} performs intermediate computations (\eg, differences, ratios, aggregations) over extracted data, and \texttt{ChartMoE.describe} generates explanatory textual analyses that connect the visual structure, quantitative reasoning steps, and final answer.
\end{itemize}

\noindent \textbf{Diagram Formalization.}
This layer converts visual diagrams into symbolic structures amenable to automated reasoning. \emph{DiagramFormalizer} parses geometric and schematic diagrams into ConsCDL and ImgCDL‑style formal languages, enabling symbolic representations for downstream deductive geometric reasoning.
\begin{itemize}
    \item \textbf{CDL.}
The \emph{CDL} tools convert geometric diagrams and problem statements into formal constraint languages suitable for symbolic solvers. 
\texttt{image\_cdl} maps raw diagram images to ImgCDL-style primitives (points, lines, circles, incidences), whereas \texttt{text\_cdl} parses textual problem descriptions into ConsCDL-style constraints and goals. 
\texttt{construction\_cdl} recovers the underlying construction sequence of a diagram, and \texttt{goal\_cdl} extracts explicit target predicates (\eg, lengths, angles, congruence relations), enabling theorem-driven reasoning in downstream geometry solvers.  
    \item \textbf{Symbolic Parsing.}
The \emph{SymbolicParsing} tools build fully symbolic problem representations by combining diagram parsing with formal language construction. 
\texttt{Inter-GPS} parses both the problem text and the associated diagram into a unified formal language and then applies theorem-based symbolic reasoning step by step to solve geometry problems. 
\texttt{DiagramFormalizer} provides complementary parsing endpoints that expose ImgCDL/ConsCDL representations directly, supplying structured inputs for interpretable geometry problem solvers such as Inter-GPS.  
\end{itemize}

\noindent \textbf{Math Solvers.}
The solver layer performs domain‑specific high‑level reasoning on top of structured perceptual inputs:
\begin{itemize}
    \item \textbf{Multimodal}: \texttt{G‑LLaVA} is a domain-specific model that integrates a CLIP-based vision encoder with DeepSeekMath-RL to bridge the gap between visual perception and complex mathematical reasoning. It executes geometry‑aware multimodal reasoning by aligning figure interpretation with textual reasoning through specialized instruction tuning.
    \item \textbf{Math}: \texttt{MultiMath} is a specialized multimodal model for geometric problem solving, optimized through geometric cross-modal alignment and instruction tuning to accurately interpret figures and relationships. It bridges visual perception and symbolic mathematics by combining a CLIP‑based vision encoder with DeepSeekMath‑RL, tackling complex multimodal math problems end‑to‑end.
\end{itemize}

%% file: table/tab-data.tex
\begin{table}[t]
\centering
\renewcommand\arraystretch{0.93}
\caption{Dataset statistics in \env{}.}
\fontsize{8}{10}\selectfont\setlength{\tabcolsep}{0.3em}
\resizebox{1.0\linewidth}{!}{ %
\begin{tabular}{l @{\hskip2.5pt} | c | c | c }
\toprule
\textbf{Dataset} & \textbf{Type} & \textbf{\# of training instances} & \textbf{\# of testing instances}\\
\midrule
FigureQA & ID & 2000 & -- \\
ChartQA & ID & 1500 & 1250 \\
Geometry3k & ID& 2000& 600 \\
GeoQA & ID & 2000 & 500 \\
UniGeo & ID& 2000 & 750 \\
MathVision & ID & 2000 & -- \\
MathV360 & ID & 1500 & -- \\
MapQA & ID & 1500 & 2000 \\
InfographicVQA & ID & 1000 & -- \\
ThinkVL & ID & 2000 & -- \\
VizWiz & ID & 200 & -- \\
ScienceQA & ID & 1000 & -- \\
DocVQA & ID & 970 & -- \\
CLEVR & ID & 1000 & -- \\
A-OKVQA & ID & 1500 & -- \\
\midrule
TABMWP & OOD & -- & 970 \\
AI2D & OOD & -- & 810 \\
PlotQA & OOD & -- &2000 \\
Clever-math & OOD & -- & 930\\
IconQA & OOD & -- & 1500\\
Mathvista & OOD & -- & 1000\\
\midrule
\textbf{Total} & -- &\textbf{ 22,170}&\textbf{12,310}\\
\bottomrule
\end{tabular}
}
\label{tab:datastats}
\end{table}

%% file: appendix/app-baseline.tex
We include additional details of baseline tool-integrated or reasoning-augmented VLMs as follows:
\begin{itemize} 
    \item \textbf{VTool-R1}~\cite{wu2025vtool} integrates multimodal tool invocation directly into the VLM reasoning loop via reinforcement learning. By optimizing for feedback signals during problem-solving, it surpasses standard prompting and supervised fine-tuning baselines in complex visual reasoning tasks. 
    \item \textbf{Perception-R1}~\cite{yu2025perception} utilizes RL to learn active perception policies that optimize how VLMs attend to and interpret visual inputs. This approach moves beyond passive image encoding, fostering decision-oriented visual understanding through reward-driven refinement. 
    \item \textbf{R1-VL}~\cite{zhang2025r1} employs GRPO to enhance step-by-step visual reasoning. By incentivizing effective intermediate reasoning steps rather than solely focusing on outcomes, it aligns visual perception with logical inference throughout the generated trajectory. 
    \item \textbf{R1-OneVision}~\cite{yang2025r1} establishes a framework for converting visual inputs into structured, cross-modal formal representations. By leveraging these symbolic intermediates during reinforcement learning, the model achieves generalized reasoning capabilities across heterogeneous task types. 
    \item \textbf{LLaVA-OneVision}~\cite{an2025llava} proposes a unified training paradigm that consolidates diverse visual tasks under a consistent interface. This design enables a single general-purpose VLM to generalize across wide-ranging visual domains without requiring task-specific adaptation.
\end{itemize}

%% file: appendix/app-prompt.tex
We include the prompt template in Figure~\ref{fig:prompt}.

\begin{figure}[htbp]
\centering
\begin{tcolorbox}[
    colback=gray!15,
    colframe=gray!75,
    fonttitle=\large\bfseries\sffamily\color{white},
    coltitle=white,
    bottomrule=0pt,
    toprule=0pt,
    leftrule=0pt,
    rightrule=0pt,
    rounded corners,
]
\textbf{System Prompt:} \\[2pt]
You are an advanced AI agent capable of complex reasoning and tool usage. You must strictly adhere to the following protocol for every interaction: \\
1. ALWAYS call the appropriate tool first; \\
2. NEVER provide answers without tool results; \\
3. Call appropriate tools based on the task; \\
\textcolor{blue}{\{tool\_descriptions\}} \\
4. Reasoning Before Action: before selecting a tool, you must analyze the user's request and determine the necessary steps. Output your internal monologue and logic inside \texttt{<think>} and \texttt{</think>} tags; \\
5. Tool Execution: If a tool is required, generate the tool call immediately after your reasoning. Use the following standard JSON format wrapped in XML tags:
\texttt{<tool\_call>\{"tool": "<tool\_name>", "task": "<task\_name>"\}</tool\_call>} \\
6. Reasoning After Action: Once you receive the output from a tool, you must analyze the results to determine if further actions are needed or if the task is complete. Output this analysis inside \texttt{<think>} and \texttt{</think>} tags;\\
7. Final Output: When you have formulated your conclusion, you must wrap your final answer in \texttt{<answer>} and \texttt{</answer>} tags.

\medskip
\textbf{User Prompt:} \\[2pt]
Please answer the following \textcolor{blue}{\{task\_type\}} question: \\
Question: \textcolor{blue}{\{Question\}} \\
Please provide a complete step-by-step solution to this problem. Your reasoning should: \\
1. Analyze the problem systematically \\
2. Check if the tool execution and answer are correct \\
3. If there are errors, explain what went wrong and provide the correct reasoning \\
4. Provide the final answer \\
Use natural expressions like 'let me think' or 'hmm' when helpful, but keep it concise. It’s encouraged to use self-reflection or verification especially in the verifying tool output in the reasoning process. \\
Provide your detailed reasoning between \texttt{<think>} and \texttt{</think>} tags, then give your final answer between \texttt{<answer>} and \texttt{</answer>} tags. \\
Output format: \textcolor{blue}{\{output\_format\}}
\end{tcolorbox}
\caption{Prompt template.}
\label{fig:prompt}
\end{figure}

%% file: appendix/app-method.tex
\section{Additional Method Details of \ours{}}
\noindent \textbf{Image Additional Design.} 
Integrating the InternVL3~\cite{zhu2025internvl3} family into our reinforcement learning framework requires a bespoke adaptation of the visual processing pipeline, as its composite architecture differs substantially from monolithic models such as Qwen2.5-VL~\cite{bai2025qwen2}. Instead of performing early tensor conversion, we preserve raw \texttt{<image>} placeholders in the prompt and rely on InternVL's native processor to expand them into 256 \texttt{<IMG\_CONTEXT>} tokens and inject the corresponding visual embeddings during the forward pass. To ensure stability in distributed training, we customize Ray and Fully Sharded Data Parallel (FSDP) to wrap only the language decoder layers, leaving the vision encoder unsharded for memory efficiency, and we adjust attention masks and position IDs in VLLM to accommodate the extended visual token sequence. This minimally invasive interface adaptation preserves InternVL's native visual processing while enabling robust, efficient multimodal GRPO training.

%% file: appendix/app-implmenetation.tex
\subsection{Detailed Error Types for Error Analysis}
\label{app:error}
We define the detailed error types as follows:
\begin{itemize}
    \item \textbf{E1: Invocation schema violation (wrong function-call structure).} The model produces an invalid tool call that violates the prescribed schema, such as missing required fields, extraneous keys, incorrect nesting, or non-JSON-conformant structures that prevent execution.
    \item \textbf{E2: Invalid argument name (wrong argument key).} The tool call structure is syntactically valid, but one or more argument \emph{names} do not match the tool specification (\eg, using \texttt{"x\_axis"} instead of \texttt{"x\_label"}), causing the call to be rejected by schema validation.
    \item \textbf{E3: Invalid argument value (wrong argument format).} Argument names are correct, but the \emph{value types or formats} are invalid, such as providing strings where numbers are required, out-of-range values, or malformed lists that violate the tool’s input constraints.
    \item \textbf{E4: Incorrect argument value (wrong argument content).} The tool call is syntactically valid and passes type checks, but the \emph{semantic content} of one or more arguments is wrong (\eg, selecting the wrong region, axis, or series for analysis), leading the tool to operate on an incorrect target.
    \item \textbf{E5: Invalid output from tool execution (wrong answer format).} The tool executes but returns an output that does not conform to the expected format for downstream use, such as missing required fields, malformed JSON, or values that cannot be parsed into the canonical answer representation.
    \item \textbf{E6: Incorrect reasoning from tool execution.} The tool output is valid and informative, but the model fails to map it to the correct final answer, due to faulty logical deductions, misinterpretation of intermediate results, or inconsistent multi-step reasoning.
\end{itemize}

\subsection{Additional Training Setup Details}
We configure the training environment with a maximum of three turns per interaction, balancing exploration depth and computational cost. The agent retains full access to the conversation history and tool interaction logs across turns, and up to $24$ trajectories are processed concurrently to maximize throughput.
Agent-side code execution is implemented in Python~3.10.
Training uses $8$ NVIDIA H200 GPUs with the FSDP2 (Fully Sharded Data Parallel) strategy for efficient distributed optimization. We set parameters to temperature $0.7$. The pipeline integrates a tool-router service for real-time tool execution and feedback, and employs asynchronous rollouts with a batch size of $128$ to maintain high GPU utilization.
We consider four different backbones for \ours{} with varying sizes, including InternVL3-2B/8B/14B and Qwen2.5-VL-7B, in the main experiment.
We perform ablation experiments and additional analysis with InternVL3-8B as the VLM.